\documentclass[10pt,twocolumn,letterpaper]{article}

\usepackage[pagenumbers]{iccv} 

\definecolor{iccvblue}{rgb}{0.21,0.49,0.74}
\usepackage[pagebackref,breaklinks,colorlinks,allcolors=iccvblue]{hyperref}

\usepackage{bm}
\usepackage{todonotes}
\usepackage{multicol}
\usepackage{multirow}
\usepackage{diagbox}
\usepackage{comment}
\usepackage{color}

\title{RatBodyFormer: Rat Body Surface from Keypoints}

\author{Ayaka Higami$^{1}$\quad Karin Oshima$^{2}$ \quad Tomoyo Isoguchi Shiramatsu$^{2}$ \\
\quad Hirokazu Takahashi$^{2}$ \quad Shohei Nobuhara$^{3}$ \quad Ko Nishino$^{1}$\\
\\
\vspace{\baselineskip}
$^{1}$ Graduate School of Informatics, Kyoto University  \qquad  \\
$^{2}$ Graduate School of Information Science and Technology,
The University of Tokyo \qquad \\
$^{3}$ Faculty of Information and Human Sciences, Kyoto Institute of Technology \\
\texttt{\small \href{https://vision.ist.i.kyoto-u.ac.jp/research/ratbodyformer/}{https://vision.ist.i.kyoto-u.ac.jp/research/ratbodyformer/}}
}

\begin{document}

\twocolumn[{
  \maketitle 
  \begin{center}
    \captionsetup{type=figure}

    \includegraphics[width=\linewidth]{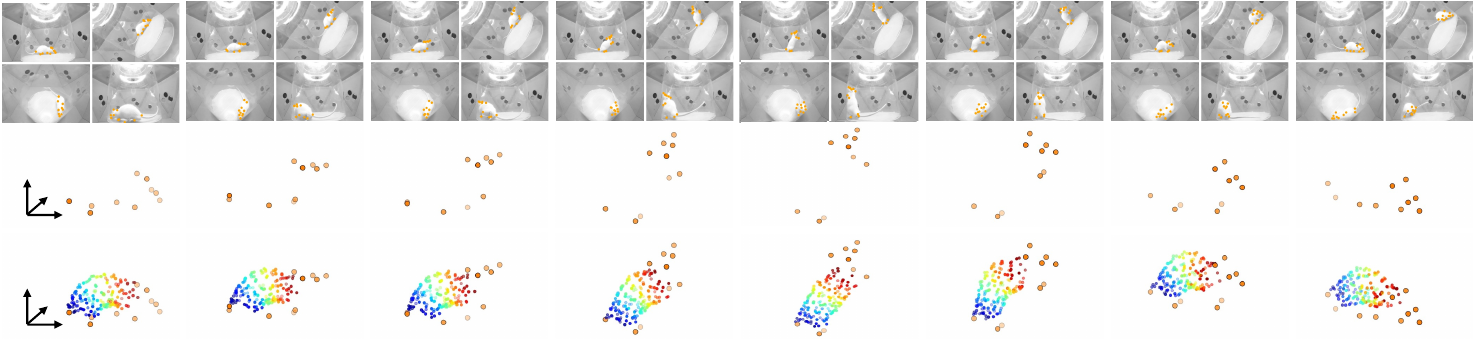}
    \vspace{-0.5\baselineskip}
    \captionof{figure}{
     We introduce a novel multiview camera system (RatDome) to capture multiview videos of rats (top), and a novel Transfomer-based network (RatBodyFormer) that recovers the deforming body surface as a dense set of surface points (bottom: rainbow points) predicted from detectable keypoints (middle: orange points). The body surface reconstructed with RatBodyFormer offers a much richer window into the complex rat behavior.} 
    \label{fig:opening figure}
\end{center}
}]

\begin{abstract}

Analyzing rat behavior lies at the heart of many scientific studies. Past methods for automated rodent modeling have focused on 3D pose estimation from keypoints, \eg, face and appendages. The pose, however, does not capture the rich body surface movement encoding the subtle rat behaviors like curling and stretching. The body surface lacks features that can be visually defined, evading these established keypoint-based methods. In this paper, we introduce the first method for reconstructing the rat body surface as a dense set of points by learning to predict it from the sparse keypoints that can be detected with past methods. Our method consists of two key contributions. The first is RatDome, a novel multi-camera system for rat behavior capture, and a large-scale dataset captured with it that consists of pairs of 3D keypoints and 3D body surface points. The second is RatBodyFormer, a novel network to transform detected keypoints to 3D body surface points. RatBodyFormer is agnostic to the exact locations of the 3D body surface points in the training data and is trained with masked-learning. We experimentally validate our framework with a number of real-world experiments. Our results collectively serve as a novel foundation for automated rat behavior analysis.

\end{abstract}

\section{Introduction}
\label{sec:intro}

Rodent behavior analysis underpins the scientific process of many areas in biomedical and neuroscientific research. 
The behavioral outcomes of rodents play a key role in validating hypotheses that lead to scientific discoveries. Careful observations of rodent behaviors, however, incur costs in manpower and time, causing a bottleneck in the scientific process. Human observation is also prone to errors. Subtle behavioral differences may be missed or misjudged~\cite{Automated_imagebased,learning_to_recognize}. If we can automate this experimental validation process while ensuring its accuracy or even extend it beyond the human level, we may dramatically expedite the scientific discovery process~\cite{A_robust_automated_system,Continuous_Whole_Body_3D,computational_analysis,Computational_Neuroethology}. 

We focus on rats. Rats are the main animal model for neuroscientific experiments, due to the complexity of their behavioral repertoire and their body size, which is suitable for measurement systems such as electrophysiology~\cite{mice_vs_rat_1,mice_vs_rat_2,mice_vs_rat_3}.
A growing number of recent works, indeed, target computational rat behavior analysis or provide the basis for it. 
Maghsoudi~\etal~\cite{superpixcel} painted markers on the body and tracked them based on their colors.
Matsumoto~\etal~\cite{3d_rat_recovery} tracked rat body parts by fitting a physical body model to the point clouds. 
More recently, Mathis~\etal~\cite{DLC} introduced a pre-trained deep network that can be fine-tuned to extract 2D keypoints of interest. This provides a general framework for skeleton estimation of animals, very much in the spirit of human pose estimation~\cite{Deeper_cut}. Following the work of Mathis~\etal, many works have achieved deep learning-based pose estimation of rats in 2D~\cite{AMBER,ye2024superanimal} and in 3D~\cite{dannce,Freipose2020} from videos or images. 

These past methods, however, are pose estimation and only provide us with the means to understand rat behavior through the movements of sparse keypoints that are well-defined visually. With DeepLabCut~\cite{DLC}, these keypoints for a rat would typically lie on the face, hands, feet, and tail. Although the movements of these keypoints may suffice for a range of applications,  for instance, to recognize when a rat is feeding itself, they only give an extremely sparse sampling of the entire body which tells us much less than what the detailed movements of the body may convey.

The largest missed opportunity lies in the body surface, the fur-coated elastic body that shows subtle twitches, twists, curl-up, stretching, and even hair standing which all eloquently speak to the rich inner conditions of the rat and colorize their behavior with context.
Accurate reconstruction of the body surface is difficult with a keypoint-based method, even with large-scale pre-training, as the rat body lacks any 
visual features whatsoever for such keypoint detectors to latch on to. It is also highly non-rigid and deformable, much more so than the human body. As such simple interpolation of detectable keypoints cannot work. 

One may imagine a few approaches for reconstructing this featureless, non-rigid body surface. For chimpanzees, Sanakoyeu~\etal~\cite{DensePose_chimps} successfully leveraged DensePose~\cite{densepose} by transferring the surface points on the chimpanzee body onto the canonical uv-textured surface. This reuse of DensePose, however, is only possible for proximal animal classes of humans as they have the same body structure. For other classes of animals, we are forced to restart from training DensePose itself, which necessitates massive pixel-to-surface annotations. This is not only impractical but infeasible as there is no principled means to define a UV coordinate on the featureless and highly deformable rat body. Instead, Bola\~{n}os~\etal~\cite{virtual_mouse} used a 3D virtual mouse mesh model by CT scanning an anesthetized mouse. MAMMAL~\cite{MAMMAL} tracks the body of a mouse by deforming this same mouse mesh to align with detected keypoints and silhouettes of multi-view images. The body of an awake rat is, however, completely different from an anesthetized one, let alone a rat is very different from a mouse, and cannot be CT-scanned. 

How then can we accurately model the 3D body surface deformation of rats? Our goal is to model the non-rigid surface deformation and realize full 3D rat reconstruction including the body surface without interfering with its natural behavior. 
We achieve this by learning to predict a dense set of body surface points from the detectable facial and appendage keypoints. In other words, our method learns the coordination between the body surface and other body parts namely the head, feet, and tail.

We make two key technical contributions for this. The first is a novel means to collect training data. We need a sizable amount of training data that associates representative keypoints including facial features and appendages with densely sampled points of the body surface. The rat body surface is featureless and deformable, to the point that it is near impossible even for a human to annotate. We overcome this by temporarily attaching trackable points (colorful beads) and passively capturing them in a novel multi-camera system which we refer to as the RatDome. We show that we can automatically recover the keypoints together with the 3D body surface points at each time instance with multiview geometry. We call this first-of-its-kind dataset, the RatDome dataset. The second is a novel network to transform detected 3D keypoints to 3D body surface points. We realize this with a transformer-based model that is agnostic to the exact locations of the 3D body surface points in the training data and introduce masked-learning to train it. We refer to this model as RatBodyFormer.

We experimentally validate our framework with a number of real-world experiments using rats for neuroscientific studies. We first show that RatBodyFormer can estimate the body surface accurately regardless of their poses or shapes. Our model can estimate their body surface with an average L2 error of around 6.5 mm, approximately the same diameter as the beads (5.5 mm) attached to the body surface. We also achieved lower errors of body surface points compared to  MAMMAL~\cite{MAMMAL}.
We demonstrate the use of the recovered body surface by constructing an animatable rat model using Gaussian Splatting~\cite{3Dgaussiansplatting}, which we refer to as GaussianRat. GaussianRat can be rendered in arbitrary pose driven by the keypoints, which can be leveraged for analysis-by-synthesis in downstream applications. 
We believe our RatDome dataset and RatBodyFormer offer foundational tools for advancing the automation of rat behavior analysis and together open a new avenue of research towards computer vision for science.

\section{Related Work}\label{sec:related_works}

\paragraph{Rodent Behavior Analysis}

Many methods have been proposed for automatic rodent behavior modeling~\cite{mocap_rat, superpixcel,long_term_tracking_socialstructure}. Mimica~\etal~\cite{mocap_rat} reproduced human mocap on rats with retroreflective markers. Maghsoudi~\etal~\cite{superpixcel} painted markers on the rat body and tracked them based on their colors. These methods are invasive and can lead to significantly altered behaviors as they require markers on the rat body in the actual experiments. 

Markerless, non-invasive methods have also been proposed. Belongee~\etal~\cite{smart_vivarium} proposed the particle filter based algorithm to track contours of mice. Chaumont~\etal~\cite{mice_tracking} tracked mice by modeling the body parts using geometric primitives and linked them with physical constraints.  
These methods do not recover the body surface itself.


Recently, deep learning based methods~\cite{LEEP,lightningpose,DLC, Freipose2020,dannce,liftpose3d,ye2024superanimal,single_3D, deepposekit, AMBER} have become popular for detecting keypoints. DeepLabCut~\cite{DLC} detects user-defined keypoints from monocular images based on the network of 2D human pose estimation~\cite{Deeper_cut}. DANNCE~\cite{dannce} and FreiPose~\cite{Freipose2020} integrate multiview images to estimate 3D keypoints. LiftPose3D~\cite{liftpose3d} achieves monocular 3D animal pose estimation by leveraging a deep neural network for lifting 2D human poses to 3D~\cite{humanliftpose3d}. 
All of these methods can only reliably detect sparse keypoints (~\eg facial points, paws, midline) with well-defined visual features, or interpolate them on the bone which is rigid. The body surface, however, cannot be localized by these methods because it has no easily identifiable feature of color or shape and as it is highly non-rigid. We reconstruct the body surface passively without any markers by modeling the relationship of detectable keypoint positions and the body surface deformation.


A few recent works have targeted rodent body surface modeling~\cite{Armo,MAMMAL, virtual_mouse}. Bola\~{n}os~\etal~\cite{virtual_mouse} constructed a 3D virtual mouse mesh model from CT data. To generate synthetic training data for pose estimation, they deform the scanned anesthetized body surface by articulating the bones. MAMMAL~\cite{MAMMAL} tracked a mouse body surface and extremities in 3D by fitting the same mouse mesh model~\cite{virtual_mouse} so that it aligns with the detected 2D keypoints and the silhouettes in multiview images. 
The deformation of the anesthetized mouse surface, however, cannot accurately model the surface deformation of an awake rat. 
We capture the body surface of an awake rat by attaching beads and capturing it from multiple views, and learn to predict those coordinates from detectable keypoints.


\vspace{-8pt}
\paragraph{3D Animal Reconstruction}

One common approach for 3D human reconstruction is to learn a statistical model from pre-acquired 3D scans to better condition the solution space~\cite{SMPL,SMAL,VAREN, FLAME:SiggraphAsia2017}. SMPL~\cite{SMPL} is a statistical 3D human linear blended shape model parameterized by the person's shape and pose. The pose parameters are the joint rotations and the shape parameters are the PCA coefficients of a large collection of aligned body scans. SMAL~\cite{SMAL} extends this to quadrupled animals. Most 3D reconstruction methods regress these parameters directly from the image with a neural network~\cite{HMR, Youwang2021Unified3M} or optimize their parameters so that projected keypoints align with their imaged ones~\cite{smplify, MuvS, SPIN,DMMR}. 
There is no such statistical model of rats, mainly because there is no large-scale collection of 3D scans aligned with a canonical 3D surface.
A straightforward approach for building such 3D surface dataset is to capture the 3D surface of the target, \ie, rats, first, and align them \wrt a canonical 3D scan~\cite{dynamicfaust,non-rigid-registration-survey}. This approach, however, cannot provide reliable alignments because most of the points on the rat body surface cannot be localized by their 3D shape or surface color.

A few works directly regress the 3D body surface from images~\cite{animalavatars,DensePose_chimps,densepose}. AnimalAvatars~\cite{animalavatars} regresses continuous surface embedding~\cite{CSE} from an image to associate each pixel with a point on the target 3D shape in its canonical pose. This CSE estimation, however, requires large-scale dense annotation between 2D image pixels and a 3D canonical surface ala DensePose~\cite{densepose}, which is costly and unreliable for featureless body surfaces, or requires the target animal to be in a proximal class of humans~\cite{DensePose_chimps}. This annotation process is not only labor-intensive but also limits the accuracy of the 3D surface estimation since the mapping between each pixel and the surface point is determined intuitively by human annotators. 
We instead leverage beads and multiview geometry to learn mapping from sparse keypoints to dense body surface points, which ensures robust positional consistency. 
We use beads instead of fluorescent dye~\cite{color_label_rat}, since they are easier to distinguish by color.

\section{Method}
\label{sec:method}
We reconstruct the 3D body surface of a freely moving rat by learning the mapping from the 3D positions of visually well-defined keypoints, \ie, face and appendages, to the 3D coordinates of a dense set of body surface points. 
First, we develop a novel multiview camera system which we refer to as the RatDome (\cref{sec:rat_surface_dataset}) and collect a first-of-its-kind dataset that associates densely sampled body surface points with detectable keypoints (RatDome Dataset). Second, we derive a transformer-based network, RatBodyFormer, that takes the keypoint 3D coordinates as input and outputs sampled body surface point 3D coordinates (\cref{sec:deformable_mesh_model}). The model is trained on the RatDome Dataset. 

The experimental protocol of this study received approval by the Committee on the Ethics of Animal Experiments at the Graduate School of Information Science and Technology at the University of Tokyo (Permit Number: JA23-6).

\subsection{RatDome Dataset}\label{sec:rat_surface_dataset}

Regardless of the pre-training, a network can only learn to detect points that exhibit sufficient visual features. As a result, even after fine-tuning, an off-the-shelf deep keypoint detector \cite{DLC} can only reliably detect points on the face, appendages, and midline of the rat. Our goal is to learn to extrapolate the body surface from these keypoints. For this, we will need a sufficiently large-scale dataset of paired sets of 3D keypoints and 3D body surface points. This is challenging as the rat body is completely featureless.

\begin{figure}[t]
  \centering
  \includegraphics[width=\linewidth]{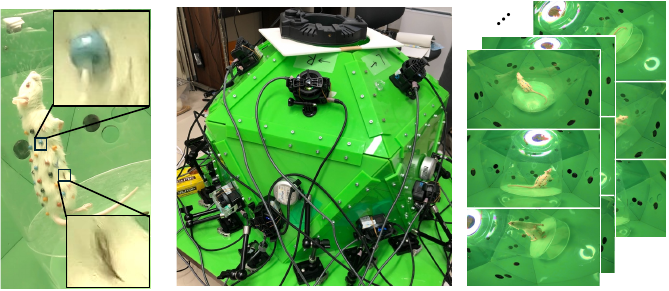}
   \caption{
   A color-beaded rat (left) and RatDome (right).
   We attach color (red, black, orange, blue) beads  and paint on the rat body surface . RatDome is a novel multiview camera studio for freely moving rats. It is shaped as a 15-faced gyroelongated pentagonal pyramid. With 15 cameras and their multiview geometry, we collect, annotate, and reconstruct paired sets of 3D keypoints and 3D body surface points of rats of different ages in weeks.
   }
  \label{fig:beadsrat_and_ratome}
  \vspace{-8pt}
\end{figure}

We overcome this challenge by simply endowing the rat body with visual features. As shown in 
\cref{fig:beadsrat_and_ratome}, 
we attach color beads to the body as well as paint markers in the areas where beads cannot be attached. We refer to these two types of body markers, beads and paints, simply as \textit{markers}. It is important to note that these markers are only used for training data capture and the rats are completely in their natural form when their behavior is observed in actual experiments. These markers are also small enough that they do not alter the rat's behavior as far as we could tell.

We build a novel multi-camera system to passively observe the color-beaded rat and reconstruct the 3D coordinates of individual markers. As shown in 
\cref{fig:beadsrat_and_ratome}, 
this RatDome is a rat-scale multiview studio. Its shape is a gyroelongated pentagonal pyramid with 15 faces. Each face is an equilateral triangle with sides of 400mm, and accommodates up to three cameras or microphones on it while also serving as a green background of the system. For our data capture, we mount one camera on each face totaling 15 views. RatDome follows the modular design of CMU Panoptic studio~\cite{panopticstudio} and can adapt to new capture devices by replacing each face.

By capturing freely moving rats of different ages (7, 9, 11 week-old) in RatDome, we collect multiview videos and the paired sets of 3D keypoints and 3D markers. We recorded 2 sessions for the 7 week-old, 1 session for the 9 week-old, and 2 sessions for the 11 week-old. Each session was approximately 10 to 15 minutes. We manually annotate the marker IDs for the 7 and 11 week-old rats for about 1100 frames each. We semi-automatically annotate (see \cref{sec:semi_annotation_pipeline}) the 7, 9, and 11 week-old rats' marker IDs for about 9000, 1800, and 15000 frames, respectively. The 7, 9, 11 week-old rats have 58, 33, and 66 surface markers, respectively. We refer to this first-of-its-kind large-scale rat body surface dataset as the \textit{RatDome Dataset}.

A notable difference of the RatDome Dataset from other datasets used for building parametric 3D shape models~\cite{SMPL,VAREN,SMAL} is that the marker positions across different individuals are not consistent as it is impossible to make them the same. Although the marker positions are the same within a single rat for multiple capture sessions, it is not strictly aligned for different rats. In other words, our RatDome Dataset can be seen as a collection of multiple motion captures each of which has slightly different annotations, similarly to the SuperAnimal dataset~\cite{ye2024superanimal}. This means that we cannot just apply PCA to obtain a universal statistical model~\cite{SMPL,VAREN,SMAL}.

\subsection{RatBodyFormer}\label{sec:deformable_mesh_model}

The markers in RatDome Dataset are not consistent across individual rats and the body surface deforms non-rigidly depending on the pose. To model this highly nonlinear deformation with inconsistent annotations, we derive a novel transformer-based model (RatBodyFormer) to recover dense 3D body surface points from sparse keypoints.

\vspace{-8pt}
\paragraph{Model Formulation}

RatBodyFormer takes $K$ 3D keypoint coordinates as input and outputs $N$ 3D surface point coordinates.  The keypoints are chosen as points locatable in images, and we use the following 10 points: \textit{nose, right and left eyes, ears, front paws, back paws, and the base of tail}. 

\begin{figure*}[t]
  \centering
  \includegraphics[width=\linewidth]{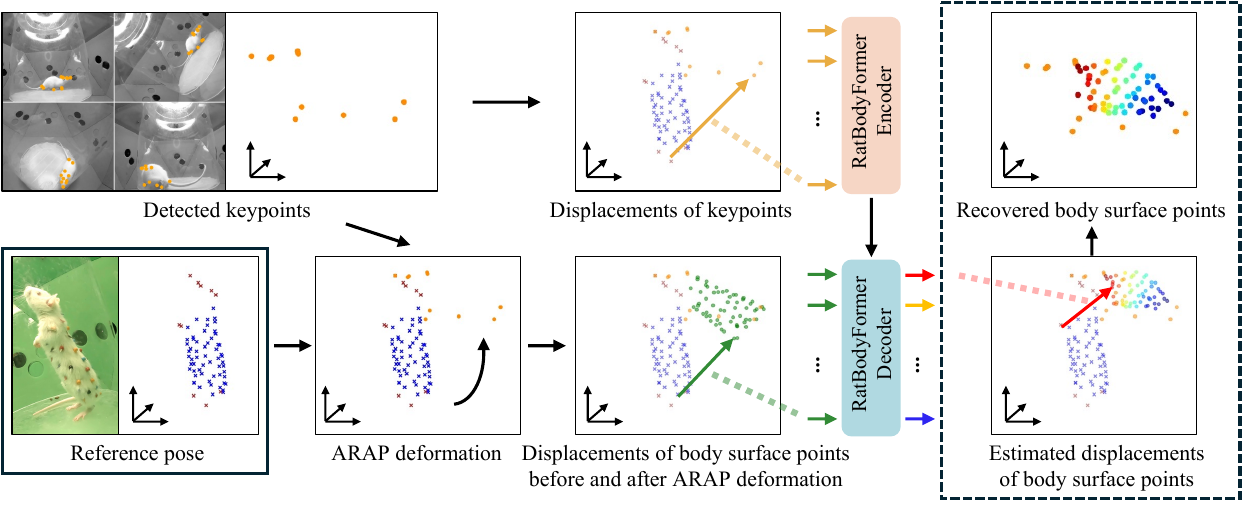}
  \caption{RatBodyFormer is an encoder-decoder Transformer model that takes the normalized displacements of detected 3D keypoints and outputs the normalized displacements of densely sampled 3D body surface points. The displacements are \wrt the reference pose. }
  \label{fig:draft_meshmodel}
  \vspace{-8pt}
\end{figure*}

As shown in \cref{fig:draft_meshmodel}, RatBodyFormer is designed as a Transformer encoder-decoder model~\cite{transformer}. Each encoder input token represents each keypoint, and each decoder output token represents each body surface point. 

\vspace{-8pt}
\paragraph{Canonical 3D Body Surface}
To consolidate annotations based on different marker positions, RatBodyFormer employs a canonical body surface $\tilde{S}$ onto which all markers are mapped. 
We pre-select a reference pose that appears multiple times in the captured sequences. By using the 3D keypoint positions as the deformation constraint, we can align all the individually observed 3D surfaces for that pose with ARAP deformation~\cite{arap}. This brings all marker positions for this reference pose to a single surface. 

In our RatDome Dataset, we manually selected a standing-on-two-feet pose as shown in \cref{fig:beadsrat_and_ratome} as the reference pose such that almost all of the whole body surface is visible from the cameras.
Please refer to the appendix for details. As a result, the keypoints and the surface points are associated with 3D points on $\tilde{S}$ as $\tilde{\mathbf{P}} = \{\tilde{p}_i\}_1^K$ and $\tilde{\mathbf{B}} = \{\tilde{b}_j\}_1^N$, respectively.  $\tilde{\mathbf{P}}$ and $\tilde{\mathbf{B}}$ are constant regardless of the body shape and pose.

\vspace{-8pt}
\paragraph{Network Architecture}
To represent different body shapes of different rats, we introduce point-wise scaling and translation parameters for individual rats, $\mathbf{C}_\mathrm{P} = \{c_{p_i}\}_1^K, \mathbf{T}_\mathrm{P} = \{t_{p_i}\}_1^K$ and $\mathbf{C}_\mathrm{B} = \{c_{b_j}\}_1^N, \mathbf{T}_\mathrm{B} = \{t_{b_j}\}_1^N$ for keypoints and body surface points, respectively. These scaling and translation parameters are optimized in the training process. At inference time for a rat whose scaling and translation parameters are unknown, we test-time optimize the parameters while keeping RatBodyFormer frozen.

The keypoints and the body surface points are first rotated around the Z-axis, \ie, the direction of gravity, such that the rat face is oriented to a pre-determined direction. Their 3D coordinates are then normalized to $[-1,1]$: $\mathbf{P} = \{p_i\}_1^K$ and $\mathbf{B} = \{b_j\}_1^N$. The same normalization is also applied to $\tilde{S}$.

As shown in \cref{fig:draft_meshmodel}, the decoder outputs are the estimated normalized displacements of the body surface points from their coordinates in the reference pose. The decoder queries are normalized displacements of the body surface points in an initial guess $\hat{\mathbf{B}}$ from their coordinates in the reference pose. The initial guess is calculated analytically. 
We use ARAP~\cite{arap} and deform $\tilde{S}$ with the constraint that $\tilde{\mathbf{P}}$ is aligned with $\mathbf{P}/\mathbf{C}_\mathrm{P}$. As a result, we obtain the ARAP-deformed $\tilde{\mathbf{B}}$ as $\hat{\mathbf{B}} = \{\hat{b}_j\}_1^N$. For each keypoint and body surface point, we use the displacement vector normalized by the identity-dependent parameters $p_i/ c_{p_i} - (\tilde{p}_i + t_{p_i}) $ and $\hat{b}_j - (\tilde{b}_j + t_{b_j}) $ as the inputs to the model, respectively.

The encoder and the decoder first 
embed each of these displacements in higher dimension $d$ together with positional encoding.
The positional encoding for keypoints and body surface points is defined with a set of sinusoidal functions of the projection of their positions $\tilde{p}_i$ and $\tilde{b}_j$ on $\tilde{S}$ by the Laplace-Beltrami eigenfunctions defined on the canonical body surface $\tilde{S}$~\cite{CSE,levy2006laplace}.
The outputs of the RatBodyFormer decoder are $N$ normalized displacement vectors $\{\delta_j\}_1^N$. Each $\delta_j$ reconstructs the body surface coordinates as $c_j (\delta_j + \tilde{b}_j + t_{b_j})$. 
We set $d = 128$ in the experiment.

\vspace{-8pt}
\paragraph{Loss Functions}\label{sec:pseudo-label}
The primary loss function for training RatBodyFormer is the supervision provided by the annotations in the RatDome Dataset. We directly compare the estimated 3D body surface point coordinates and their ground-truth coordinates with the L2 norm as $\mathcal{L}_\mathrm{3D}$. 
This L2 norm loss $\mathcal{L}_\mathrm{3D}$, however, can be defined only for a subset of the $N$ body surface points. Since they are defined as the union of all annotated body surface points on different rat surfaces, a single input keypoint set from a rat has a ground-truth annotation only for the body surface points of the same rat. 
To ensure that the body surface point positions are consistent, we introduce a position consistency loss $\mathcal{L}_\mathrm{p}$ when training on data of multiple rats. Please refer to the appendix for details.
In addition to $\mathcal{L}_\mathrm{3D}$ and $\mathcal{L}_\mathrm{p}$, we employ a silhouette loss $\mathcal{L}_\mathrm{s}$ which counts the number of predicted body surface points whose projections fall outside of the 2D rat region in a view~\cite{DRWR}. 


\vspace{-8pt}
\paragraph{Semi-Automatic Annotation}\label{sec:semi_annotation_pipeline}
The RatDome Dataset contains many frames capturing color-beaded rats but without manual annotations (\cref{sec:rat_surface_dataset}). We derive a semi-supervised learning method to make full use of this data. The method leverages a small number of frames fully-annotated with ground-truth to infer the labels for the remaining frames.

First, we compute the 3D coordinates of the markers. This process is automated, similar to that of the Panoptic Studio Dataset~\cite{panopticstudio}. We detect the 2D markers from each image with an object detector~\cite{yolox} trained with manual annotations. We then leverage multiview triangulation to disambiguate markers of the same color on the body surface.

Once the 3D marker coordinates are computed for each frame, we assign their marker IDs so that they are consistent with the manually annotated ones by using RatBodyFormer itself. Suppose we have trained RatBodyFormer using only the manually-annotated frames. We use this initial model to estimate the 3D coordinates of the body surface points, and find the correspondences between the triangulated 3D points and those estimated by RatBodyFormer by minimizing their Euclidean distance~\cite{hungarian}. These correspondences allow transfer of marker IDs from the manually annotated body surface points to the automatically triangulated body surface points so that RatBodyFormer can be retrained using both the manually-annotated and automatically-annotated frames.
We define these automatically-annotated labels as the \textit{semi-automatically annotated labels} to distinguish them from manually annotated labels. We experimentally show that the use of semi-automatically annotated labels improve estimation accuracy (\cref{sec:ratbody_former_accuracy}).

\vspace{-8pt}
\paragraph{Training}
The input parameters of RatBodyFormer are the keypoint coordinates $\mathbf{P}$, the individual-dependent scaling factors $\mathbf{C} = \{ \mathbf{C}_P, \mathbf{C}_B \}$, and translations $\mathbf{T} = \{ \mathbf{T}_P, \mathbf{T}_B \}$.  We alternate between the optimization of RatBodyFormer and the individual-dependent parameters. The individual-dependent $\mathbf{C}$ and $\mathbf{T}$ are initialized by $\alpha$s and $0$s, respectively. Here $\alpha$ is obtained as a result of aligning the canonical 3D body surface $\tilde{S}$ as mentioned before.
At every epoch of the training, we optimize RatBodyFormer by minimizing $\mathcal{L}_\mathrm{3D}$ and $\mathcal{L}_\mathrm{p}$, while keeping the individual-dependent parameters fixed. During this optimization, we refine $\mathbf{C}$ and $\mathbf{T}$ by optimizing $\mathcal{L}_\mathrm{s}$ and $\mathcal{L}_\mathrm{p}$ at every $T$ epochs, while keeping RatBodyFormer fixed. We use $T=50$ in our experiments.

\vspace{-8pt}

\paragraph{Inference}
We estimate the scaling and translation parameters $\mathbf{C}$ and $\mathbf{T}$ of a new rat with inference-time optimization using $\mathcal{L}_\mathrm{s}$ and $\mathcal{L}_\mathrm{p}$. The initial value $\alpha_\mathrm{p}$ for $\mathbf{C}_P$ and $\alpha_\mathrm{B}$ for $\mathbf{C}_{B}$ are manually set by the ratio of the length from the nose to the base of the tail, and the ratio of the girth length, respectively.
After optimizing $\mathbf{C}$ and $\mathbf{T}$, we regress the body surface point coordinates.


\section{Experiments}
\label{sec:experiments}

We evaluate the effectiveness of RatBodyFormer with a number of experiments each focused on validating key properties of it. In \cref{sec:ratbody_former_accuracy}, we evaluate the accuracy and generalization capability of RatBodyFormer. We also compare with MAMMAL~\cite{MAMMAL}.
%

\subsection{RatDome and RatDome Dataset}
RatDome is equipped with 15 GoPro 10 cameras and 5 Intel L515 LiDAR-Camera devices. The LiDAR cameras are installed for potential ground truth measurements of markerless rats. We found that the depth captured with the LiDAR cameras was noisy and had holes for any direct measurement. We only use the depth images for D3 of \cref{sec:ratbody_former_accuracy} by combining it with shape-from-silhouette. GoPro captures 4K videos at 60Hz, and L515 captures 1080p videos at 30Hz and 1024$\times$768 depthmaps at 30Hz. We place a translucent acrylic tube of 300 mm diameter and 5 mm thick as a rat cage to avoid the rat from moving into the corners.

All the cameras are calibrated by capturing a chessboard moved around in the dome. The mean reprojection error after regular bundle adjustment was about 1.1 pixels~\cite{hartly2004multiple}. Refraction by the acrylic tube is not modeled in the camera calibration. The cameras are temporally synchronized with flash light from a strobe. 
The flash light can potentially be observed over two frames leading to a 1/60s misalignment.

\subsection{RatBodyFormer}\label{sec:ratbody_former_accuracy}

We evaluate the accuracy of RatBodyFormer in different scenarios. We first train and test the model using a single rat where the training and the testing sets share the same marker annotations (D1). We specifically evaluate the advantage of using semi-automatically annotated labels, the frames automatically annotated with RatBodyFormer itself (\cref{sec:semi_annotation_pipeline}). Next, we combine two rats of different ages of weeks for training and testing (D2). 
This experiment also quantifies the advantage of using semi-automatically annotated labels in the case of combining two rats.
Finally, we use the entire RatDome Dataset, where we can use 7, 9, and 11 week-old annotations for training (D3), and use 5, 7, 9, 11, and 14 week-old rats without markers to evaluate the generalization capability of the proposed method.

\cref{tab:dataset} summarizes the dataset splits for these evaluations, D1, D2, and D3. We divide the manually annotated data of 7, 11 week-olds into 12 parts in a temporally sequential order, one for testing, one for validation, and the rest for training. Because our semi-automatically annotated labels sometimes mistake surface marker IDs, we use semi-automatically annotated data only for training.

\begin{table}[]
  \centering
  \scalebox{0.83}{
  \begin{tabular}{@{}ccccc@{}}
    \toprule
    Split & Test & Val & \multicolumn{2}{c}{Train}\\
      & MA &MA & MA & SAA \\ 
    \midrule
     D1 &   7w p1  & 7w p2 & 7w p3-12 & 7w \\
     D2  & 7w/11w p1  & 7w/11w p2 & 7w/11w p3-12 & 7w,11w \\
    D3 &  -  & 7w/11w p2 & 7w/11w p3-12 & 7w, 9w, 11w  \\
    \bottomrule
  \end{tabular}
}

 \caption{Our RatDome Dataset split for evaluation. ``w'' and ``p'' mean ``week-old'' and ``part'', respectively. ``MA'' and ``SAA'' are annotation types, and mean ``manually-annotated data'' and ``semi-automatically annotated data'', respectively.  }

  \label{tab:dataset}
\end{table}

\begin{figure}
    \centering
    \includegraphics[width=1\linewidth]{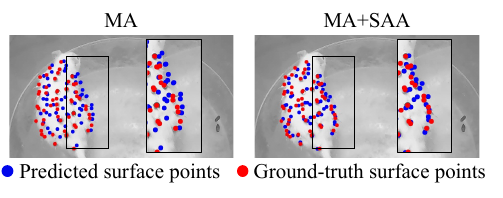}

    \caption{Qualitative results of D1. We show the results of trained by only manually-annotated data (MA) in the left, and the results of trained by manually-annotated and semi-automatically annotated data (SAA) in the right. Semi-automatically annotated data improve the body surface estimation.}
    \label{fig:eval_single}
    \vspace{-8pt}
\end{figure}

\begin{figure}[]
    \centering
    \includegraphics[width=\linewidth]{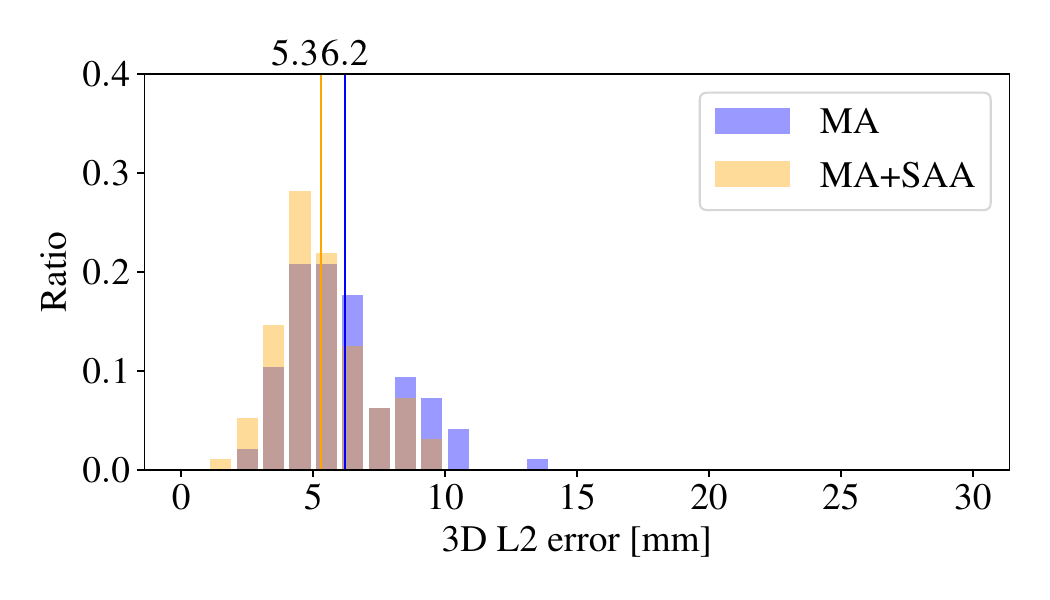}
   
    \caption{L2 error histograms of D1. Each vertical bar indicates the mean L2 error of the histogram in the same color. Our semi-automatically annotation label improve average error by about 0.9 mm.  ``MA'' and ``SAA'' mean ``manually-annotated data'', and  ``semi-automatically annotated data'',  respectively.}
    \vspace{-8pt}
 \label{fig:eval_single_histogram}
\end{figure}

\vspace{-8pt}
\paragraph{D1: Single Rat} 
\cref{fig:eval_single_histogram} shows the L2 error histograms
for D1 between the predicted body surface 3D coordinates and the corresponding manually-annotated ground truths. This result demonstrates the advantage of using semi-automatically annotated labels as they improve the accuracy by about 0.9 mm.
\cref{fig:eval_single} shows qualitative results. 

\vspace{-8pt}
\paragraph{D2: Two Rats}
The average estimation error of D2 is 7.3 mm when trained using manually annotated labels, and 6.6 mm when trained with both manually and semi-automatically annotated labels. For this case of combining two rats of different ages, the semi-automatically annotation labels also improve the accuracy. This result also shows that RatBodyFormer can estimate the body surface with an error comparable to the diameter of a bead (5.5 mm) for rats of different ages.

\begin{figure}
    \centering
    \includegraphics[width=0.95\linewidth]{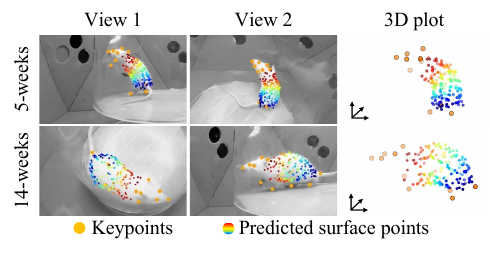}

    \caption{Qualitative results of D3. The first row and the second row show the estimated surface points for the 5 and 14 week-olds, respectively. The left and the center images show a single frame from different views. The right image shows the 3D keypoints and the 3D predicted surface points. The predicted surface points are on the rat surface.}
    \label{fig:eval_all}

\end{figure}

\vspace{-8pt}
\paragraph{D3: All Rats}
In this scenario, we train the model using 7, 9, and 11 week-old rats with the manually-annotated and semi-automatically annotated data and evaluate the accuracy on the 5, 7, 9, 11, 14 week-old rats without markers.  
For this, we define ground-truth surface points as the sum of the set of points captured by LiDAR and shape-from-silhouette. As the LiDAR points have missing values and errors, we add surface points obtained by shape-from-silhouette.
We optimize individual-dependent parameters $\mathbf{C}$ and $\mathbf{T}$ with the silhouette loss and the position consistency loss for 5 and 14 week-old rats. We obtain the mask images with SAM~\cite{SAM, SAM2}. 
The average estimation errors for 5 and 14 week-old rats are 1.80 mm and 2.40 mm, respectively. On the other hand, the average estimation errors for 7, 9, and 11 week-old rats are 2.69 mm, 1.97 mm, and 2.43 mm, respectively. Note that we calculate only the dorsal surface points because the ventral side is sometimes occluded.   
Even for individuals with body shapes not included in the training data, RatBodyFormer can estimate the body surface with an error comparable to that of individuals with body shapes present in the training data.
\cref{fig:eval_all} shows the estimated body surface points. The results show that RatBodyFormer can be applied to a variety of rats ranging from 5 to 14 week-olds. This covers the age range typically used in biomedical and neuroscientific experiments.


\vspace{-8pt}
\paragraph{Comparison with MAMMAL}
We compare the accuracy of RatbodyFormer with that of MAMMAL~\cite{MAMMAL} for the body surface marker points using D2 in \cref{tab:dataset}. 
MAMMAL uses the mouse mesh model scanned in \cite{virtual_mouse}.
As the body size of a mouse is different from a rat, we scale the virtual mouse mesh model~\cite{virtual_mouse} to match the scale of the beaded rat with a similarity transform using 8 keypoints, \ie, nose, right and left ears, front paws, back paws, and the base of the tail, in their rest poses. 
Then, we ARAP deform the surface bead points of the rat to the mouse mesh, using the 8 keypoints as its hard constraints and surface points as its soft constraints. The target position used in the soft constraint for each surface point is chosen as the mesh vertex of the mouse model closest from the surface point at each ARAP iteration.
After the alignment, we select the mesh vertices of the mouse model that are closest to the surface markers of the beaded rat as the positions of the beads in the mouse model.

\begin{figure}
    \centering
    \includegraphics[width=\linewidth]{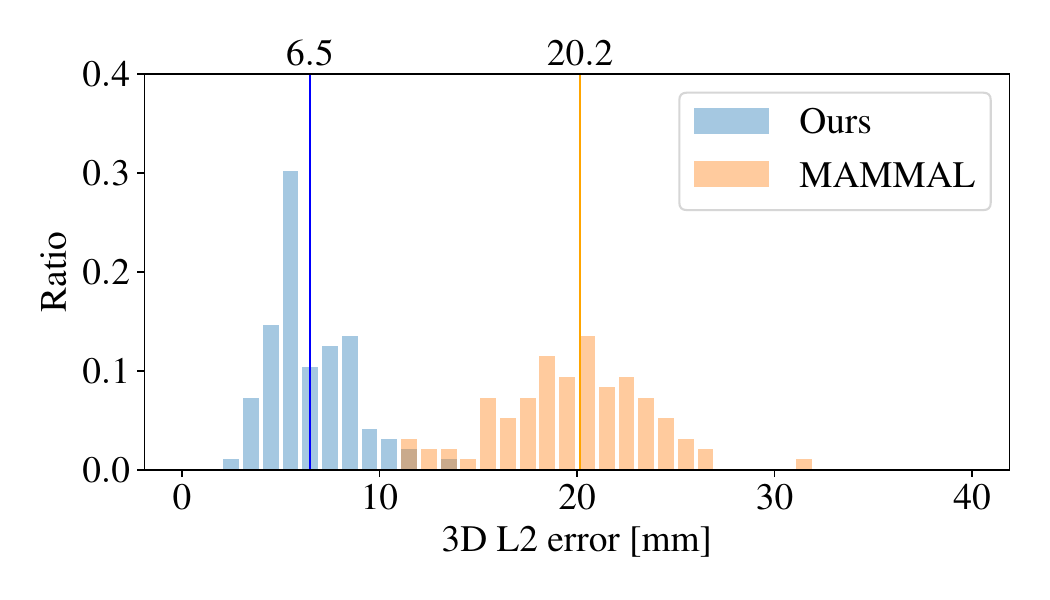}

    \caption{L2 error histogram for RatBodyFormer and MAMMAL. Each vertical bar shows the mean L2 error of each histogram in the same color. Our method achieves higher accuracy.}

    \label{fig:quauntitative_comparison_with_MAMMAL}
\end{figure}

\begin{figure}[]
    \centering
    \includegraphics[width=\linewidth]
    {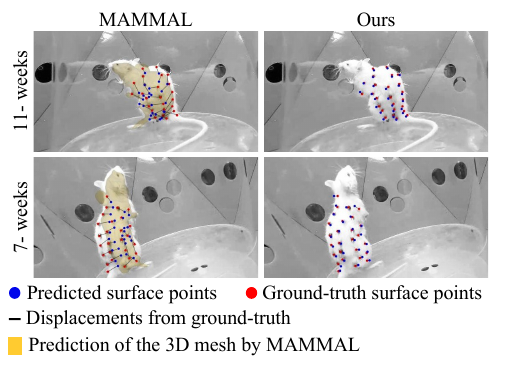}
    \vspace{-1\baselineskip}
    \caption{Qualitative comparison between RatBodyFormer (right) and MAMMAL (left).  RatBodyFormer achieves smaller displacements (black arrows) of the estimated marker positions from the ground truth.  The projections of the 3D mesh by MAMMAL visualize its inaccurate 3D shape fitting.}
    \vspace{-1\baselineskip}
    \label{fig:qualitative_comparison_with_MAMMAL}

\end{figure}

\cref{fig:quauntitative_comparison_with_MAMMAL} shows the L2 error histograms of MAMMAL and RatBodyFormer. 
RatBodyFormer achieves lower L2 errors. Note that we calculate errors only for the dorsal surface points because the ventral side is occluded in the rest pose. \cref{fig:qualitative_comparison_with_MAMMAL} shows qualitative results. Our RatBodyFormer achieves higher accuracy by a large margin.

\section{Application}

The dense body surface recovered by RatBodyFormer can enable a variety of downstream applications towards rich rat behavior analysis. Here, we demonstrate one such application of deriving an animatable 3D rat model. We refer to this model as GaussianRat as we anchor 3D Gaussian splatting to the estimated body surface. 
We define the 3D body surface mesh by using the surface points estimated by  RatBodyFormer for the torso and the points from shape-from-silhouette for the face. RatBodyFormer lets us drive the 3D body surface mesh through the placement of 3D keypoints. 

Inspired by human head avatars~\cite{psavatar,gaussianavatars, SplattingAvatar}, we attach 3D Gaussians to each mesh face and optimize their parameters. 
Each Gaussian has six properties, location $\boldsymbol{u}$, scale $s$, rotation $\boldsymbol{r}$, opacity $\sigma$, color $\boldsymbol{h}$, and attached mesh face ID $k$. 
We define each Gaussian position $\boldsymbol{u}$ by the normalized barycentric coordinates $x, y, z$  
and the pose dependent distance $l$ ($l=l^{\prime} + w \mathbf{P}$) in the normal direction of the face. Here, $l^{\prime}$, $w$, $\mathbf{P}$ denotes the pose-independent distance, weights, and keypoint 3D coordinates, respectively. We optimized the parameters except for the face ID $k$, by minimizing the L1 loss and the D-SSIM loss between the original and the reconstructed images~\cite{psavatar,gaussianavatars,SplattingAvatar}.

\begin{figure}[]
    \centering
    \includegraphics[width=0.8\linewidth]{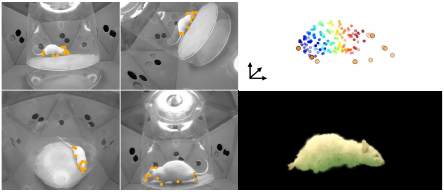}
    \caption{GaussianRat enables photo-realistic rendering of the rat body surface (bottom left) from detected keypoints (orange, left) and recovered body surface (top right).}
    \vspace{-\baselineskip}
    \label{fig:qualitative_result_of_gaussianrat}
\end{figure}

\begin{figure}[]
    \centering
    \includegraphics[width=0.95\linewidth]{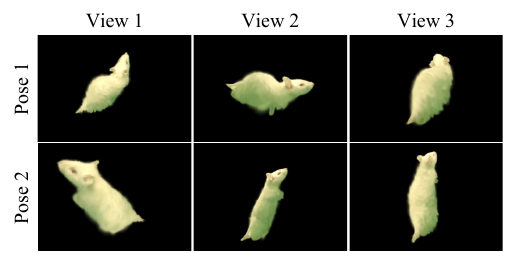}
    \caption{Frames of an animated GaussianRat.}
    \vspace{-\baselineskip}
    \label{fig:animated_gaussianrat}
\end{figure}

\cref{fig:qualitative_result_of_gaussianrat} shows the rendered GaussianRat and \cref{fig:animated_gaussianrat} shows the animated GaussianRat. Please refer to the appendix for more details. GaussianRat not only enables photo-realistic body surface estimation, but also is animatable through 3D keypoint locations. We believe GaussianRat can serve as a useful tool for analysis-by-synthesis to decipher complex rat behaviors, including those in complex interactions with other individuals and can also directly be used as virtual stimuli.

\section{Conclusion}

This paper proposed the RatDome Dataset and RatBodyFormer. The RatDome Dataset is the first-of-its-kind dataset which provides 3D body surface coordinates with temporally-consistent annotations from 7, 9, and 11 week-old rats. RatBodyFormer models the highly nonrigid deformation of the rat body and can regress the body surface 3D coordinates from its keypoint 3D coordinates. 
Experimental results demonstrate its accuracy, generalization capability, and the potential of wide application. 

Since each keypoint set corresponds uniquely to a body surface point set in the RatDome Dataset, RatBodyFormer takes only keypoints as input. But in situations not included in the RatDome Dataset, such as complex interactions, a single keypoint set may not uniquely determine a body surface point set. We plan to address this by incorporating explicit visual cues of the shape of the body surface, such as silhouettes, in our future work.
We believe that our RatDome Dataset and RatBodyFormer  collectively serve a novel, sound foundation for autonomous rat behavior analysis and will likely have far-reaching implications 
for neuroscientific research.

\section*{Acknowledgements}
This work was in part supported by
JSPS 
20H05951,
21H04893, 
23H04336, and 
24H01544, 
AMED 24wm0625401h0001, 
JST 
JPMJPR22S8, 
JPMJCR20G7, 
and JPMJAP2305, 
the Secom Science and Technology Foundation 
and RIKEN GRP.

\appendix

\setcounter{figure}{0}
\setcounter{table}{0}

\renewcommand\thefigure{\Alph{figure}}
\renewcommand\thetable{\Alph{table}}
\renewcommand{\dbltopfraction}{0.99}
\renewcommand{\textfraction}{0.01}
\renewcommand{\floatpagefraction}{0.99}
\renewcommand{\dblfloatpagefraction}{0.99}

\section{Implementation Details of Canonical Body Surface}



\begin{figure}[t]
    \centering
    \includegraphics[width=\linewidth]{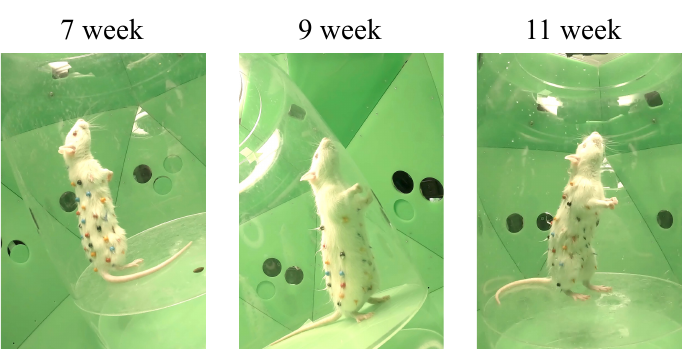}
    \caption{Manually selected standing-on-two-feet poses of 7-, 9-, 11-week-old rats. All of the markers are visible from the cameras.}
    \label{fig:reference_pose}
\end{figure}

As shown in \cref{fig:reference_pose}, we first manually select a similar pose from each individual rat to map the marker positions to a common canonical body surface. To this end, we selected a standing-on-two-feet pose in which all the markers are visible from the cameras.
Given these selected poses, we used the 7-week-old as the reference, and aligned the others by a similarity transform and ARAP deformation as follows.

We first normalized their scales, positions, and orientations by applying a similarity transform estimated from their keypoint positions.  After this similarity transform, the keypoints and surface points are aligned with ARAP deformation using the keypoints as its hard constraints and surface points as its soft constraints.  The target position used in the soft constraint for each surface point is chosen as the point on the visual hull of the target rat closest from the surface point in each ARAP iteration.



\section{Implementation Details of Position Consistency Loss}
RatBodyFormer estimates the body surface points of all annotated body surface points.  But a single input keypoint set from a rat has a ground-truth annotation only for the body surface points of the same rat. A position consistency loss $\mathcal{L}_\mathrm{p}$ is introduced to improve the position consistency of all the estimated body surface points. For each body surface point $j$, we first define three neighboring points $j_1,j_2,j_3$ in the reference pose. Then, we calculate the barycentric coordinates $(\alpha_{j_1},\alpha_{j_2},\alpha_{j_3})$ of the foot of the perpendicular dropped from $\tilde{b}_j$ to the triangle formed by $\tilde{b}_{j_1},\tilde{b}_{j_2},\tilde{b}_{j_3}$. Here, $\tilde{B} = \{~\tilde{b}_j\}_1^N$ denotes the body surface points in the reference pose. The position consistency loss $\mathcal{L}_\mathrm{p}$ is calculated as:
\begin{equation}
\mathcal{L}_\mathrm{p} = \sum_j \left\{
\begin{smallmatrix}
0 & \text{if $(\alpha_{j_1}^{\prime},\alpha_{j_2}^{\prime},\alpha_{j_3}^{\prime}>0, \alpha_{j_1}^{\prime} + \alpha_{j_2}^{\prime} + \alpha_{j_3}^{\prime}= 1 )$}, \\
\Delta_j& \text{else},
\end{smallmatrix}
\right.
\end{equation}
where $\Delta_j = \|(\alpha_{j_1}^{\prime} - \alpha_{j_1})b_{j_1}^{\prime} + (\alpha_{j_2}^{\prime} - \alpha_{j_2})b_{j_2}^{\prime} + (\alpha_{j_3}^{\prime} - \alpha_{j_3})b_{j_3}^{\prime} \|_2$, $\mathbf{B}^{\prime} = \{b^{\prime}_j\}_1^N$ is the estimated body surface points, and $(a_{j_1}^{\prime},a_{j_2}^{\prime},a_{j_3}^{\prime})$ are the barycentric coordinates of the foot of the perpendicular dropped from ${b}^{\prime}_j$ to the triangle formed by ${b}^{\prime}_{j_1},{b}^{\prime}_{j_2},{b}^{\prime}_{j_3}$. We experimentally show the effectiveness of a position consistency loss in \ref{sec:result_position_consistency_loss}.

\section{Evaluations with Other Data Splits}

To demonstrate the generalization capability of our RatBodyFormer, we further evaluate the accuracy of RatBodyFormer with other data splits.
\cref{tab:dataset_detail} shows the dataset splits of D1 and D2 scenarios. The splits ``D1'' and ``D2'' in Table 1 of the main paper appear as D1-a and D2-a in this table, respectively.

\begin{table}[t]
  \centering
  \scalebox{0.83}{
  \begin{tabular}{@{}ccccc@{}}
    \toprule
    \multicolumn{1}{c}{\multirow{2}{*}{Split}} & Test & Val & \multicolumn{2}{c}{Train}\\
     & MA &MA & MA & SAA \\ \hline
     D1-a &   7w p1  & 7w p2 & 7w p3-12 & 7w \\
     D1-b &   7w p12  & 7w p11 & 7w p1-10 & 7w \\
     D1-c &   11w p1  & 11w p2 & 11w p3-12 & 11w \\
     D1-d &   11w p12  & 11w p11 & 11w p1-10 & 11w \\ \hline
     D2-a & 7w/11w p1  & 7w/11w p2 & 7w/11w p3-12 & 7w,11w \\
     D2-b & 7w/11w p12  & 7w/11w p11 & 7w/11w p1-10 & 7w,11w \\
       \bottomrule
  \end{tabular}
}

 \caption{Our RatDome Dataset split for evaluation. ``w'' and ``p'' stand for ``week-old'' and ``part'', respectively. ``MA'' and ``SAA'' are annotation types, and stand for ``manually-annotated data'' and ``semi-automatically annotated data'', respectively.  }
  \label{tab:dataset_detail}
\end{table}

\paragraph{D1: Single Rat} 

\begin{figure}[t]
    \begin{tabular}{c}
      \begin{minipage}[]{1\hsize}
        \centering
        \includegraphics[width=\linewidth]{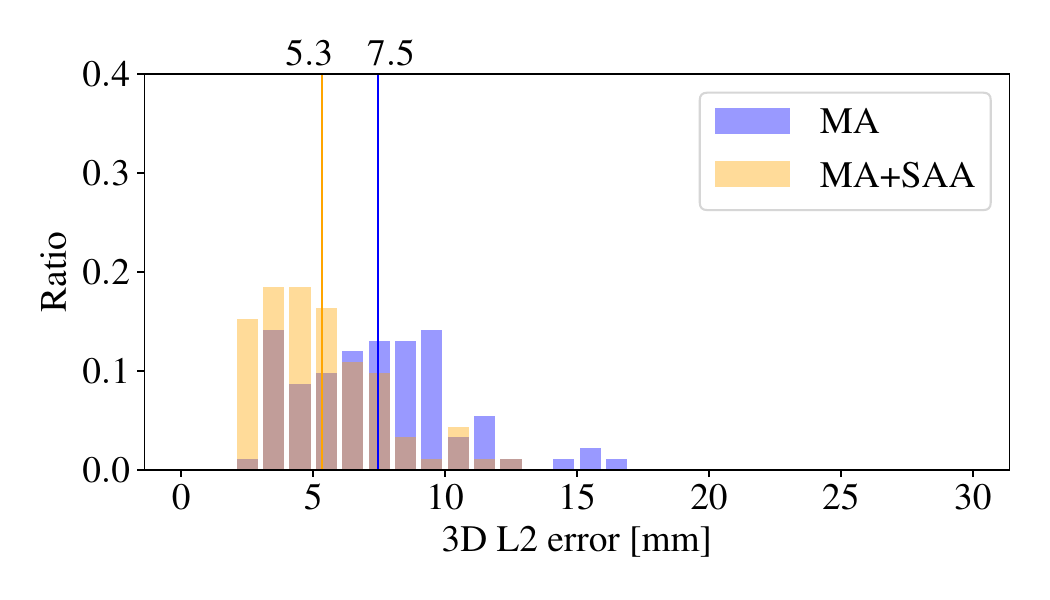}
        \subcaption{L2 error histogram of D1-b}
        \label{fig:D1_a}
      \end{minipage} \\
      \begin{minipage}[]{1\hsize}
        \centering
        \includegraphics[width=\linewidth]{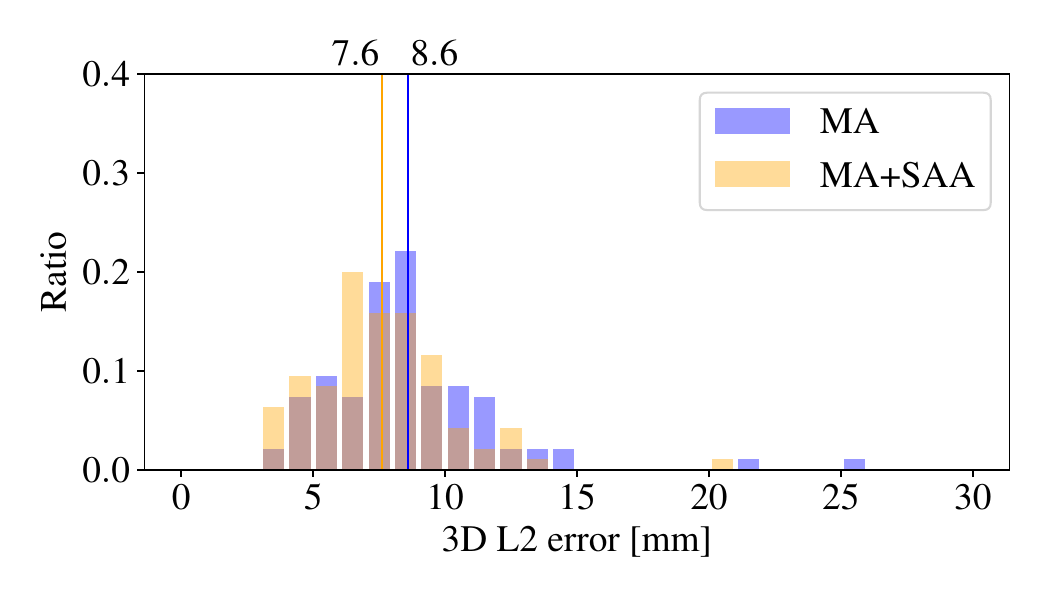}
        \subcaption{L2 error histogram of D1-c}
        \label{fig:D1_c}
      \end{minipage} \\
    \begin{minipage}[]{1\hsize}
        \centering
        \includegraphics[width=\linewidth]{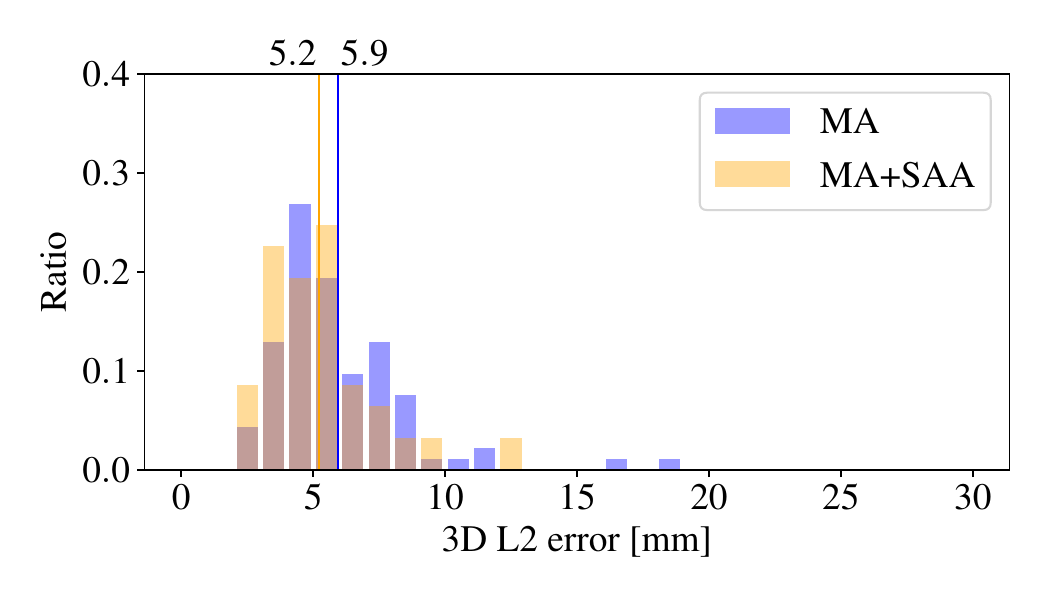}
        \subcaption{L2 error histogram of D1-d}
        \label{fig:D1_d}
      \end{minipage} \\
    \end{tabular}
    \caption{L2 error histograms of D1-b, D1-c, and D1-d. Each vertical bar indicates the mean L2 error of the histogram in the same color. ``MA'' and ``SAA'' denote ``manually-annotated data'', and  ``semi-automatically annotated data'',  respectively. Our semi-automatically annotated data improves accuracy by 0.7 mm to 2.2 mm. }
    \label{fig:D1_histgram_others}
\end{figure}

\begin{figure}
    \centering
    \includegraphics[width=\linewidth]{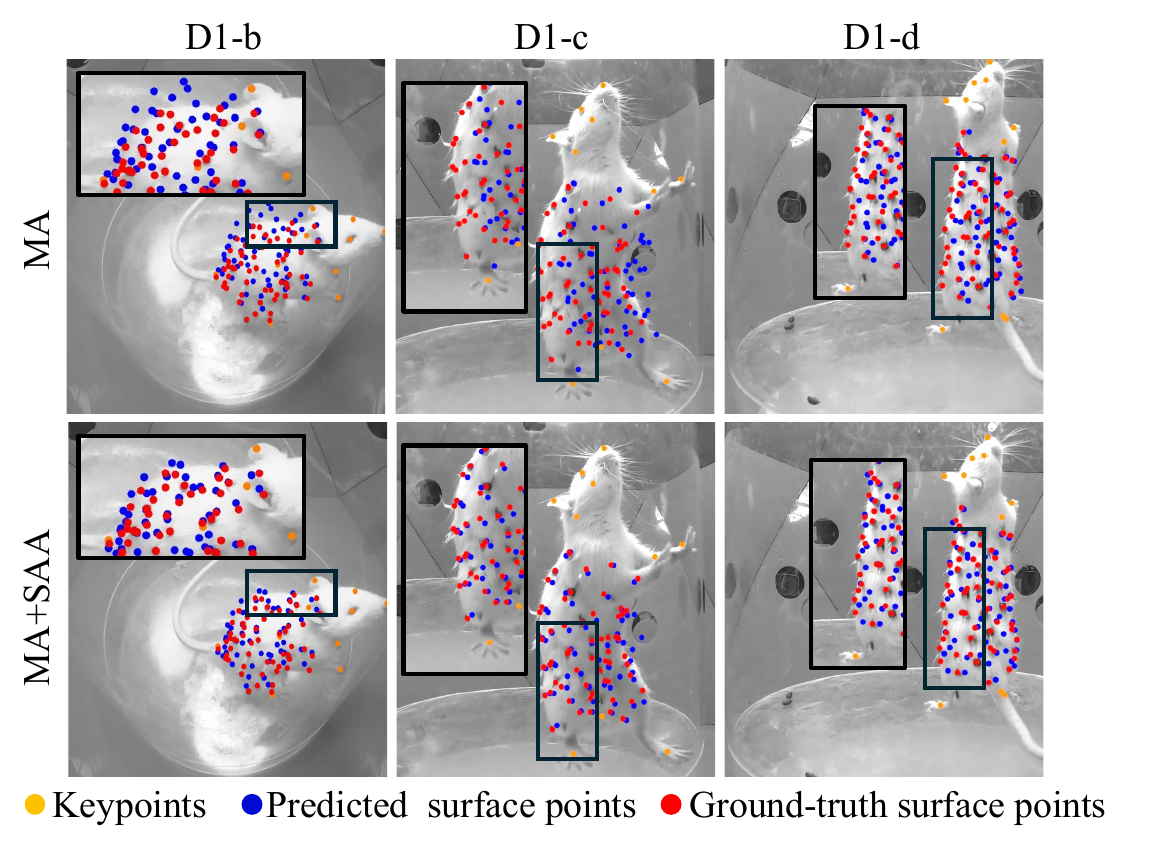}
    \caption{Qualitative results of D1-b, D1-c, and D1-d. We show the results of only using manually-annotated data (MA) for training in the top row, and the results of also using semi-automatically annotated data (SAA) for training in the bottom row. The addition of semi-automatically annotated data improves the body surface estimation.}
    \label{fig:eval_D1_another}
\end{figure}

\cref{fig:D1_histgram_others} (a), (b), and (c) show the error histograms of D1-b, D1-c, and D1-d, respectively. \cref{fig:eval_D1_another} shows qualitative results of these splits. Our semi-automatically annotated data improve accuracy in all these splits consistently.  

      

\paragraph{D2: Two Rats} 

\begin{figure}[]
    \centering
    \includegraphics[width=\linewidth]{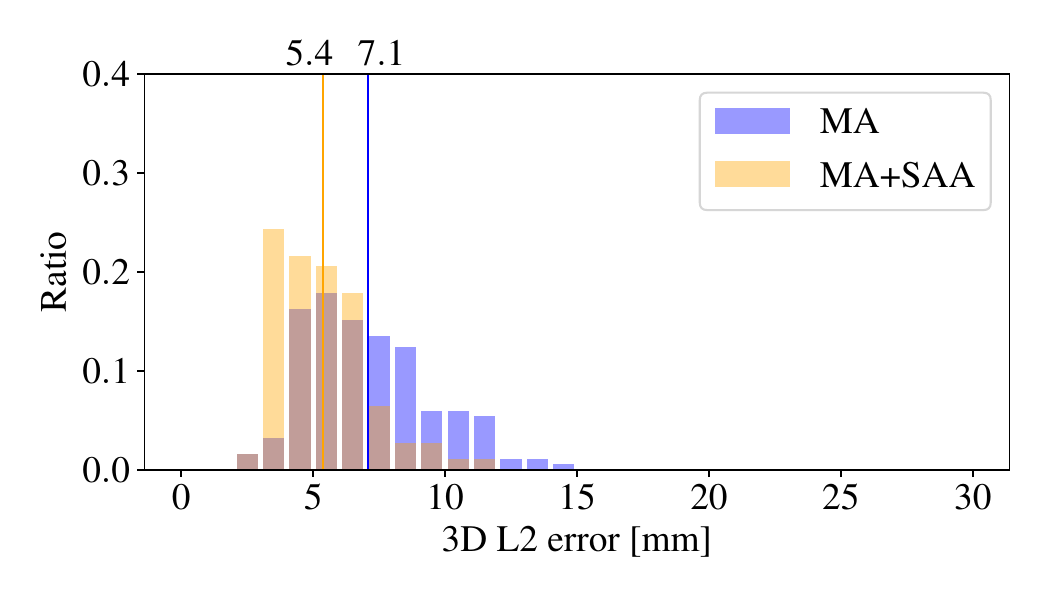}
      \vspace{-2\baselineskip}
    \caption{L2 error histogram of D2-b. Each vertical bar indicates the mean L2 error of the histogram in the same color. `MA'' and ``SAA'' represent ``manually-annotated data'', and  ``semi-automatically annotated data'',  respectively. The semi-automatically annotated data improves average error by about 1.7 mm. }
    \label{fig:D2_histgram_others}
\end{figure}

\cref{fig:D2_histgram_others} shows the error histogram of D2-b. This results demonstrate that our semi-automatically annotated data improves accuracy regardless of the dataset splits.  

\section{Evaluations of GausianRat}

\begin{figure}[]
    \centering
    \includegraphics[width=\linewidth]{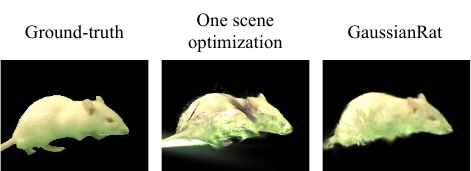}

    \caption{Qualitative results of novel-view synthesis. Compared with optimizing one scene individually, GaussianRat demonstrates photometric consistency. }
    \label{fig:qualitative_comparison_with_one_scene}
\end{figure}

In this section, we show the effectiveness of GaussianRat. 
To evaluate the photometric consistency of GaussianRat, we compare SSIM, PSNR, and LPIPS of GaussianRat to those of one scene optimization. We have synchronized 15 multiview cameras, and use 9 for testing and 6 for training. \cref{tab:photo_reality_quantitative_cimparison} shows the quantitative comparison. In \cref{fig:qualitative_comparison_with_one_scene}, we show the qualitative results of a novel view of GaussianRat and one scene optimization.  We can reconstruct the appearance consistently compared to optimizing one scene individually, when rendered from a novel view.

\begin{table}
  \centering
  \begin{tabular}{@{}cccc@{}}
    \toprule
     Method & SSIM $\uparrow$ & PSNR $\uparrow$ & LPIPS $\downarrow$ \\  \hline
     GaussianRat& $\bm{0.559}$ & $\bm{16.348}$ & $\bm{0.229}$  \\
     One scene optimization & 0.233 & 13.210 & 0.331 \\
    \bottomrule
  \end{tabular}
  \caption{Quantitative comparison. GaussianRat outperforms one scene optimization on all the metrics.}
  \label{tab:photo_reality_quantitative_cimparison}
\end{table}

\begin{figure}[]
    \centering
    \includegraphics[width=\linewidth]{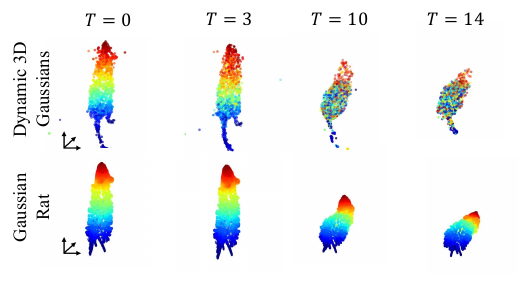}
    \caption{The Gaussian centers of Dynamic 3D Gaussians (top) and GaussianRat (bottom). The same Gaussians are represented with the same color across all frames. GaussianRat maintains consistency in the positions of the Gaussians on the body surface over time, whereas Dynamic 3D Gaussians lacks such consistency.}
    \label{fig:qualitative_geometric_consistensy}
\end{figure}

To evaluate the geometric-consistency, we compare the transition of each Gaussian center across 15 frames with Dynamic 3D Gaussians~\cite{dynamic3dgaussian}, which tracks the Gaussians over time. As shown in \ref{fig:qualitative_geometric_consistensy},  GaussianRat demonstrates higher geometric consistency.

\section{Ablation Study}

\subsection{Position Consistency Loss}\label{sec:result_position_consistency_loss}

\begin{figure}
    \centering
    \includegraphics[width=\linewidth]{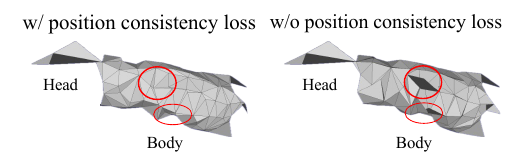}
    \caption{Qualitative results of the effect of a position consistency loss. We manually define the mesh using the body surface points in the reference pose. These body surface points are the union of the body surface points of 7- and 11-week old rats. In the case without a position consistency loss, some estimated body surface points lacks position consistency (red circle). }
    \label{fig:qualitative_comparison_of_position_coonsistency}
\end{figure}

\begin{figure}[t]
    \begin{tabular}{c}
      \begin{minipage}[]{1\hsize}
        \centering
        \includegraphics[width=\linewidth]{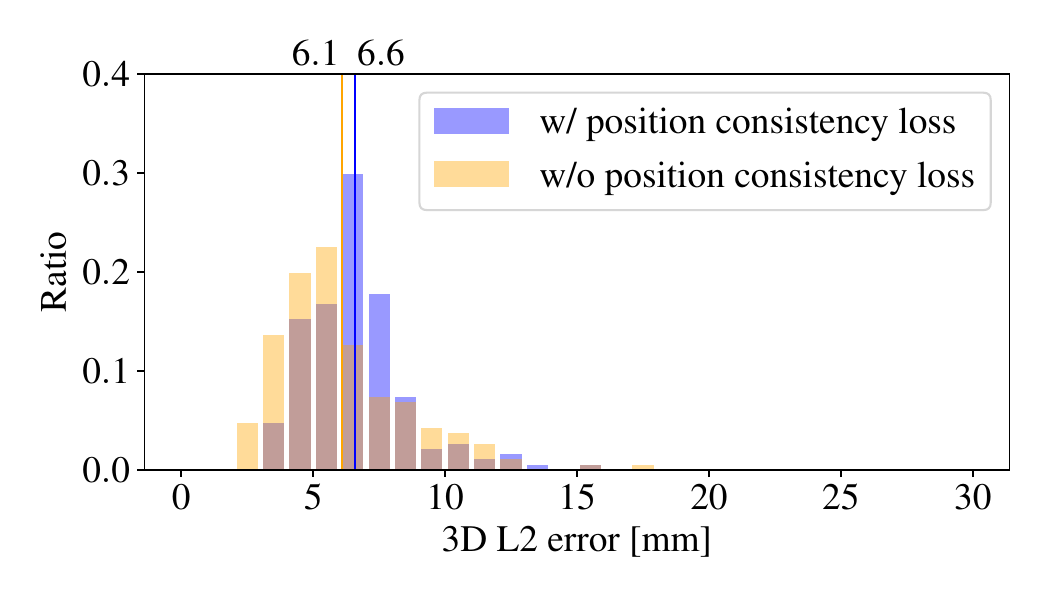}
        \subcaption{L2 error histogram of D2-a}
        \label{fig:D2_a_position}
      \end{minipage} \\
    \begin{minipage}[]{1\hsize}
        \centering
        \includegraphics[width=\linewidth]{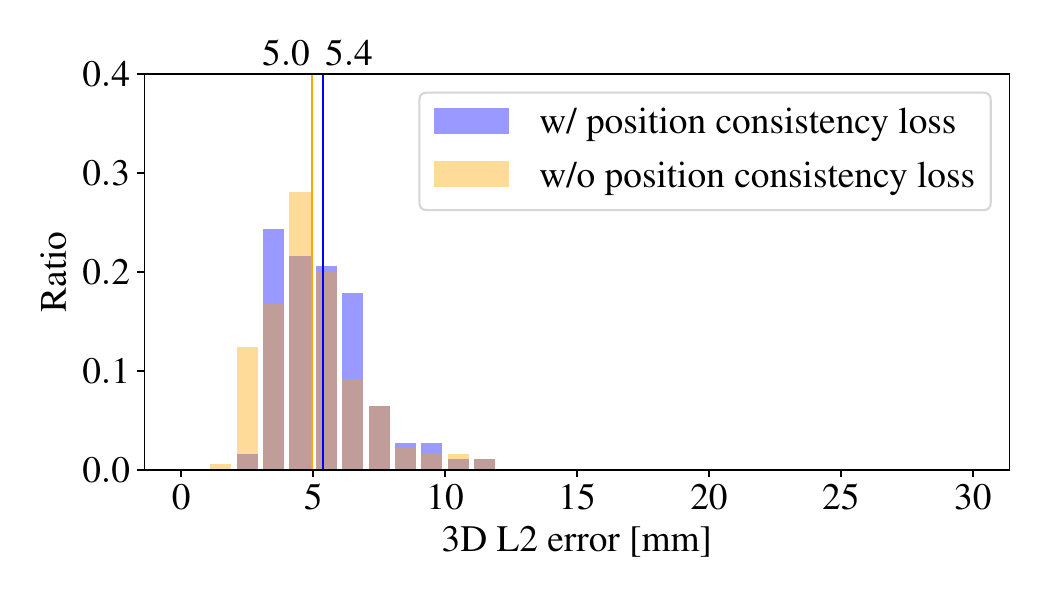}
        \subcaption{L2 error histogram of D2-b}
        \label{fig:D2_b_position}
      \end{minipage} \\
    \end{tabular}
    \caption{L2 error histograms of D2-a and D2-b. Each vertical bar indicates the mean L2 error of the histogram in the same color. The blue and orange histograms show the errors with and without a position consistency loss, respectively. Even with a position consistency loss, the estimation error of own body surface points does not significantly degrade.}
    \label{fig:D2_position_consistency_loss}
\end{figure}

This section evaluates the effect of a position consistency loss $\mathcal{L}_\mathrm{p}$ using D2-a and D2-b in \cref{tab:dataset_detail}. In this dataset, the body surface points which RatBodyFormer estimates are defined as the union of annotated body surface points on 7- and 11-week old rats.  \cref{fig:qualitative_comparison_of_position_coonsistency} qualitatively shows the effect of a position consistency loss. \cref{fig:D2_position_consistency_loss} shows L2 error histograms of estimated body surface points with and without a position consistency loss. Note that we can only calculate the error of own body surface points, and we cannot calculate it of the other individual body surface points. A position consistency loss improves the consistency of estimated body surface point positions without substantially degrading the estimation accuracy of own body surface points.

\subsection{Individual-dependent Parameters}

\begin{figure}[t]
    \begin{tabular}{c}
      \begin{minipage}[]{1\hsize}
        \centering
        \includegraphics[width=\linewidth]{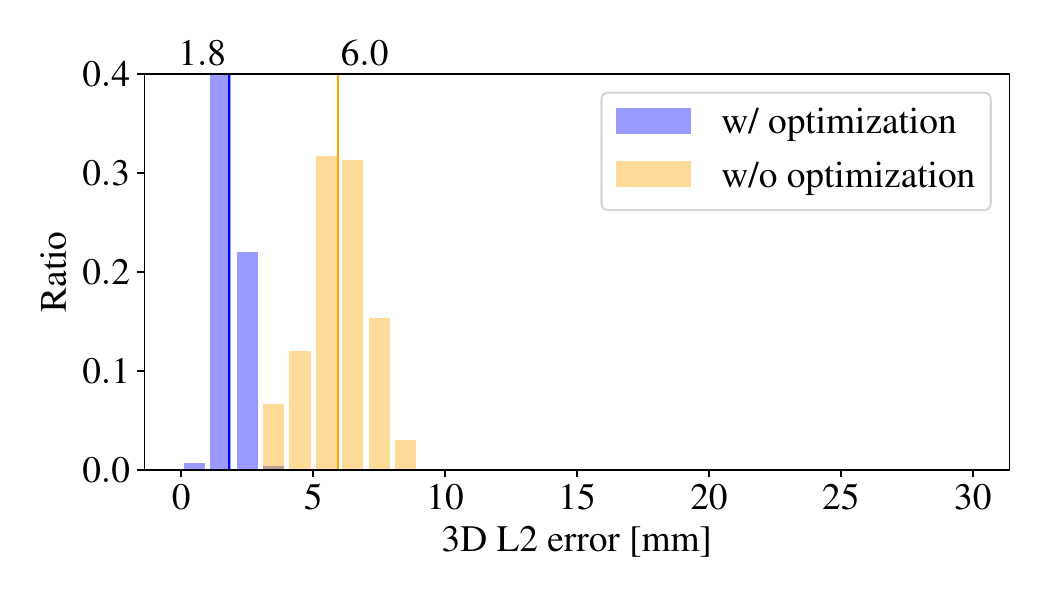}
        \subcaption{L2 error histogram of 5 week old}
      \end{minipage} \\
      \begin{minipage}[]{1\hsize}
        \centering
        \includegraphics[width=\linewidth]{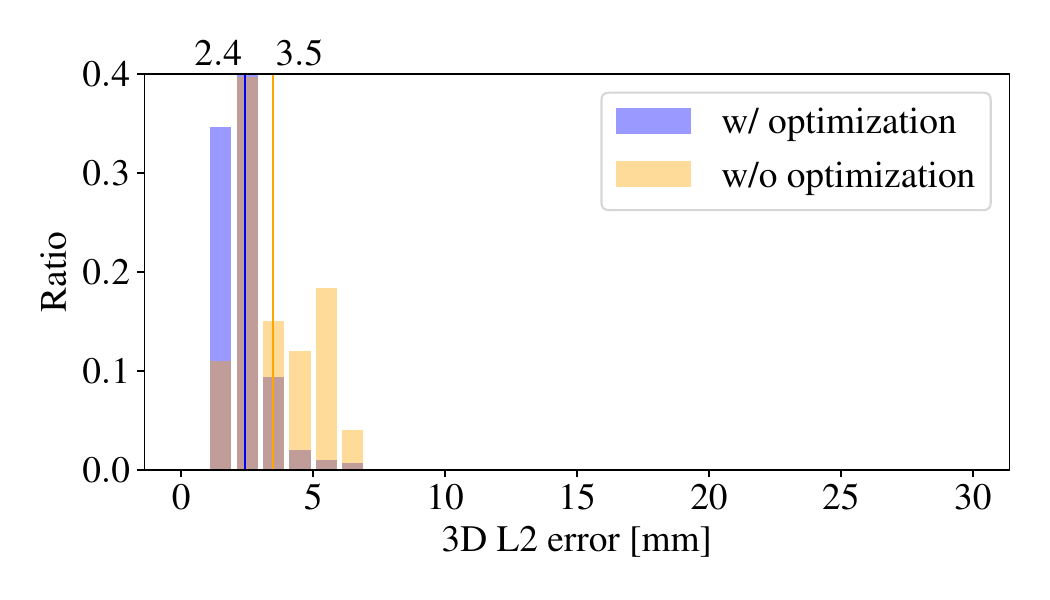}
        \subcaption{L2 error histogram of 14 week old}
      \end{minipage} \\
    \end{tabular}
    \caption{L2 error histograms of 5- and 14-week-old rats. Each vertical bar indicates the mean L2 error of the histogram in the same color. 
    The blue and orange histograms show the errors with and without optimizing the individual-dependent parameters, respectively.
    Optimizing the individual-dependent parameters consistently improves the estimation accuracy.}
    \label{fig:scalefix_D3}
\end{figure}

This section evaluates the effect of the individual-dependent parameters, \ie, point-wise scaling parameters $\mathbf{C}$ and translation parameters $\mathbf{T}$.
\cref{fig:scalefix_D3} shows the effect of optimizing the individual-dependent parameters at inference time using the D3 split in Table 1 of the main text. In this split, RatBodyFormer is trained using 7-, 9-, and 11-week-old rats with manually-annotated and semi-automatically annotated data, and evaluated with 5- and 14-week-old rats using 3D points obtained by LiDAR and shape-from-silhouette as described in the main text. For the results of inference without optimizing individual-dependent parameters, we set $\mathbf{C}= 1$ and $\mathbf{T}=0$. We can observe that the optimization improves the estimation accuracy.


\subsection{Data Normalization}

\begin{figure}[t]
    \begin{tabular}{c}
      \begin{minipage}[]{1\hsize}
        \centering
         \includegraphics[width=\linewidth]{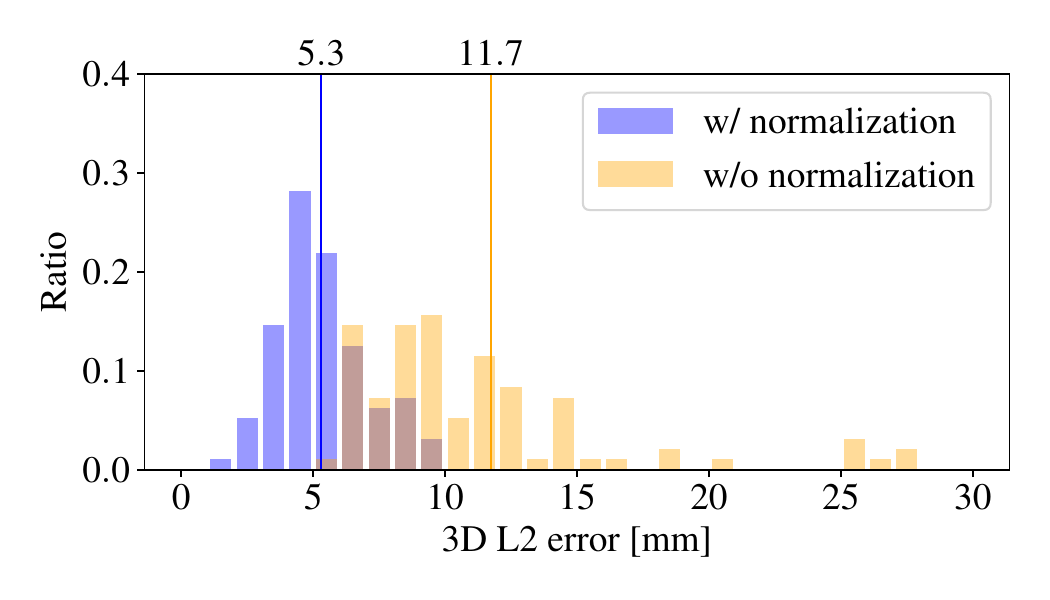}
        \subcaption{L2 error histogram of D1-a}
      \end{minipage} \\
      \begin{minipage}[]{1\hsize}
        \centering
        \includegraphics[width=\linewidth]{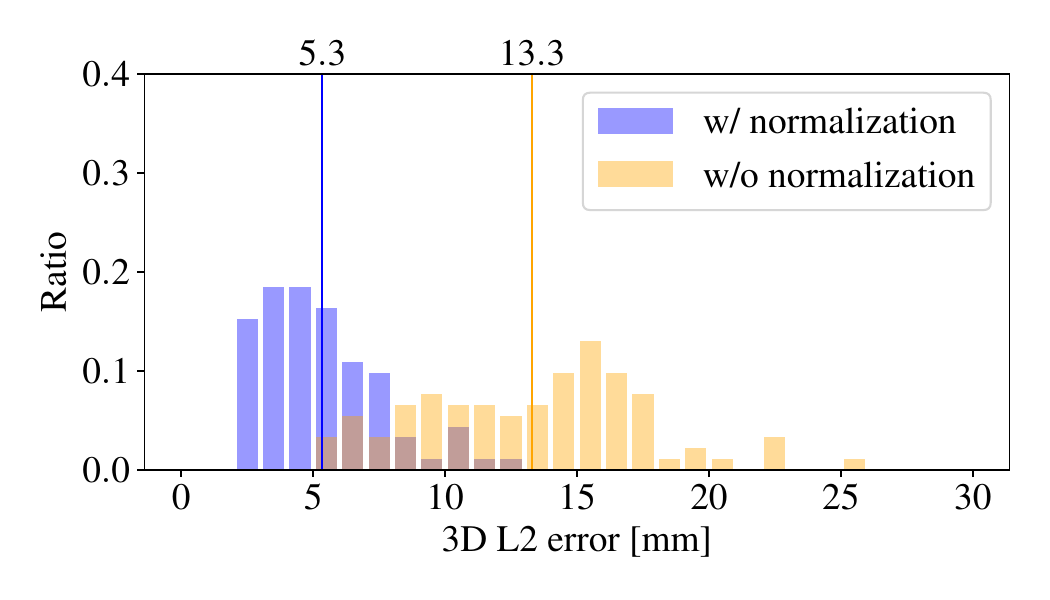}
        \subcaption{L2 error histogram of D1-b}
      \end{minipage} \\
    \begin{minipage}[]{1\hsize}
        \centering
       \includegraphics[width=\linewidth]{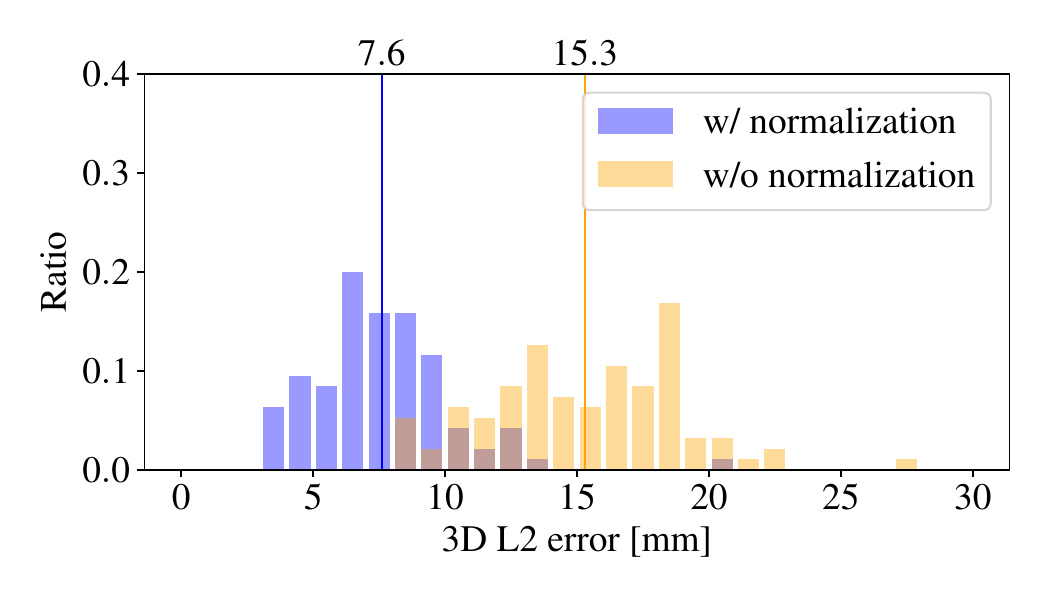}  
        \subcaption{L2 error histogram of D1-c}
        \label{fig:D1_d}
      \end{minipage} \\

    \begin{minipage}[]{1\hsize}
        \centering
         \includegraphics[width=\linewidth]{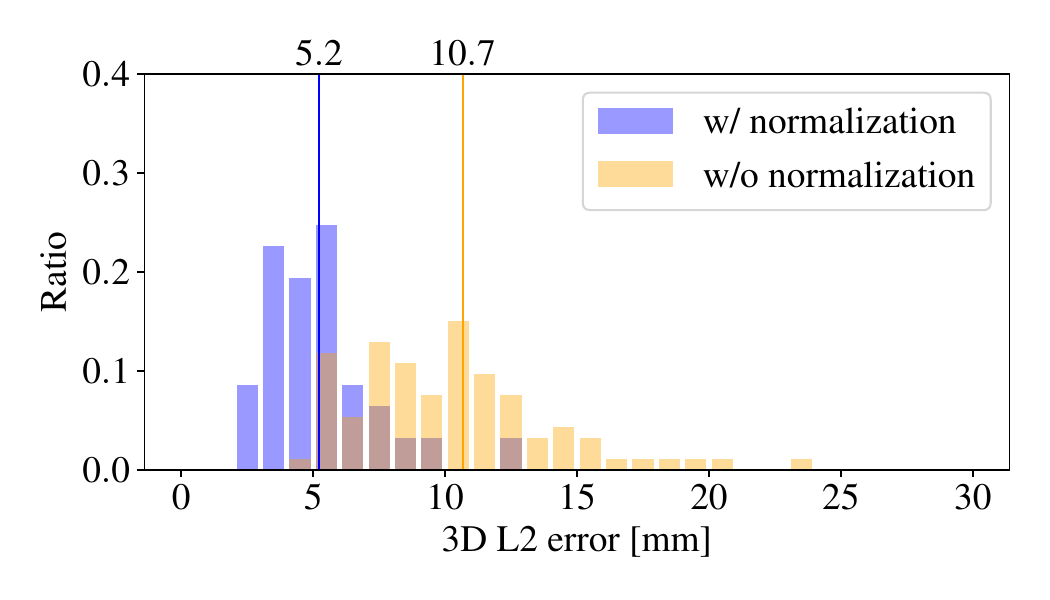}
        \subcaption{L2 error histogram of D1-d}
        \label{fig:D1_d}
    \end{minipage}
    \end{tabular}
    \caption{L2 error histograms of D1-a, D1-b, D1-c, and D1-d. Each vertical bar shows the mean L2 error of the histogram in the same color. Our data normalization improves the average L2 error by about 5.5 mm to 8.0 mm.}
    \label{fig:D1_histgram_normalization}
\end{figure}

\cref{fig:D1_histgram_normalization} shows the effect of our data normalization applied to the input of RatBodyFormer using D1 splits. 
These results clearly demonstrate that the normalization significantly improves the accuracy.


\section{Error Propagation from Keypoints}
\begin{figure}[]
    \centering
    \includegraphics[width=\linewidth]{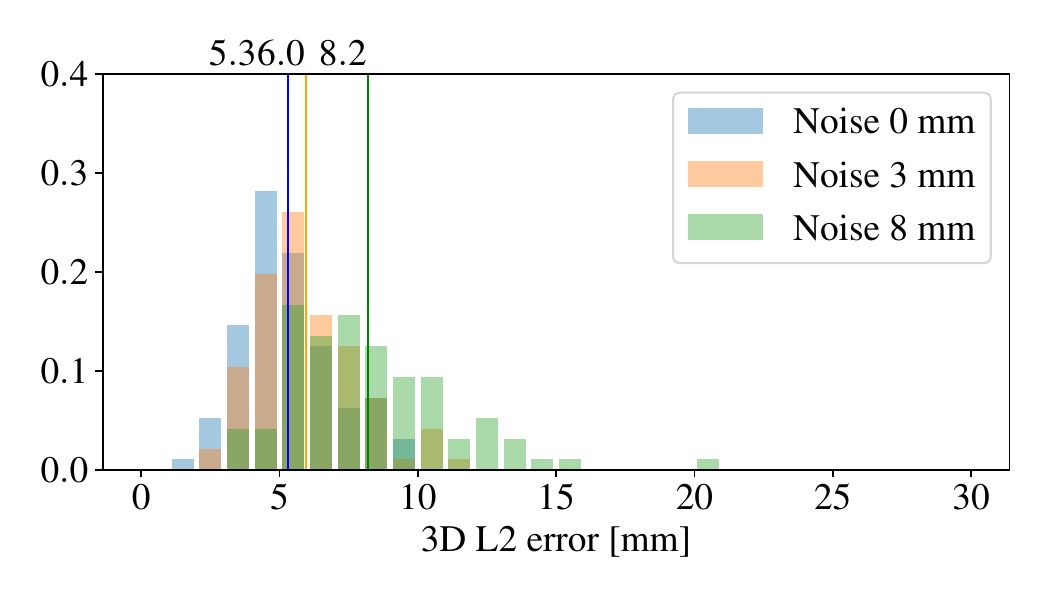}
    \caption{L2 error histograms of D1-a. Each vertical bar shows the mean L2 error of the histogram in the same color. The blue, orange and green histograms show the errors without noise, with 3 mm noise, and with 8 mm noise, respectively. Adding 8 mm noise to the input keypoints does not cause the average estimation error to increase by more than 3 mm from the case without noise. Keypoint estimation inaccuracy does not significantly affect the estimation accuracy of the body surface.
    }
    \label{fig:keypoint_estimation_dependancy}
\end{figure}

In this section,  we demonstrate the sensitivity of RatBodyFormer to the accuracy of keypoint estimation.  \cref{fig:keypoint_estimation_dependancy}
shows L2 error histograms when adding 3 mm and 8 mm noise to input 10 keypoint 3D coordinates at inference time of D1-a in \cref{tab:dataset_detail}. Even when 8 mm noise is added to the input keypoints, the average body surface estimation error increases by no more than 3 mm compared to the case without noise. This shows that the error propagation from keypoints is not significant.

\section{Initialization in RatBodyFormer}

\begin{figure}[]
    \centering
    \includegraphics[width=\linewidth]{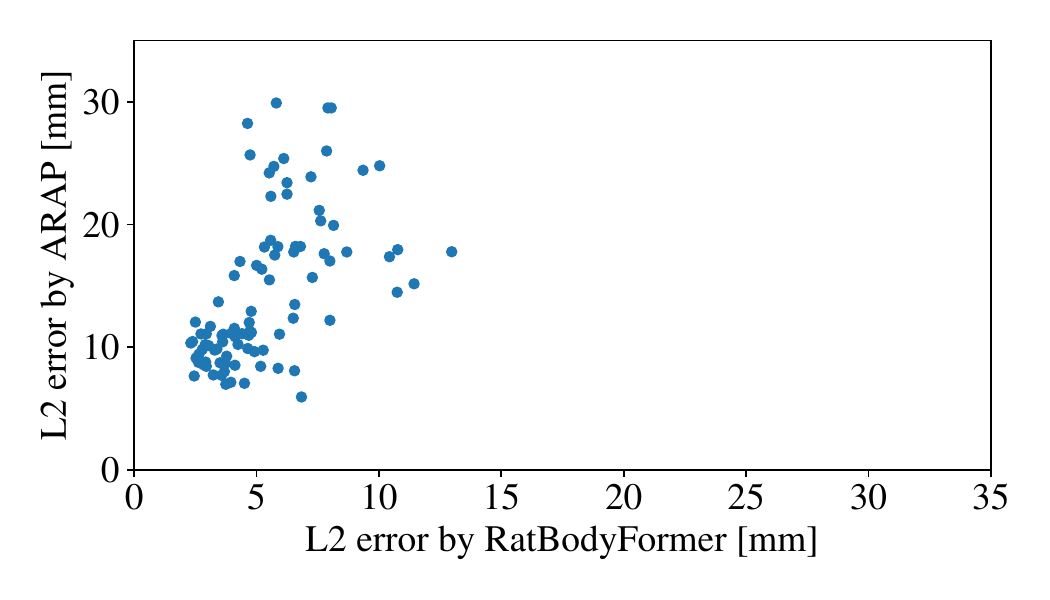}
    \caption{The relationship between the initial estimation error by ARAP deformation and the final estimation error by RatBodyFormer. RatBodyFormer's final estimation error is below 10 mm in most cases even with initial estimation errors by ARAP deformation exceeding 20 mm, which indicates its low sensitivity to initial estimation errors.
    }
\label{fig:initialization_accuracy}
\end{figure}

In this section, we demonstrate the impact of initial estimation accuracy on body surface estimation accuracy.
\cref{fig:initialization_accuracy} shows the relationship between the initial body surface estimation error by ARAP deformation and the final estimation error by RatBodyFormer. Even when the initial estimation error by ARAP deformation exceeds 20 mm, the final estimation error by RatBodyFormer is below 10 mm in most cases, indicating low sensitivity to initial estimation errors.

\clearpage

{
    \small
    \bibliographystyle{ieeenat_fullname}
    \bibliography{main}

\begin{thebibliography}{62}
\providecommand{\natexlab}[1]{#1}
\providecommand{\url}[1]{\texttt{#1}}
\expandafter\ifx\csname urlstyle\endcsname\relax
  \providecommand{\doi}[1]{doi: #1}\else
  \providecommand{\doi}{doi: \begingroup \urlstyle{rm}\Url}\fi

\bibitem[{A}lexander {M}athis et~al.(2018){A}lexander {M}athis, {P}ranav {M}amidanna, {K}evin {M}.~{C}ury, {T}aiga {A}be, {V}enkatesh {N}.~{M}urthy, {M}ackenzie~{W}eygandt {M}athis, and {M}atthias {B}ethge]{DLC}
{A}lexander {M}athis, {P}ranav {M}amidanna, {K}evin {M}.~{C}ury, {T}aiga {A}be, {V}enkatesh {N}.~{M}urthy, {M}ackenzie~{W}eygandt {M}athis, and {M}atthias {B}ethge.
\newblock {D}eep{L}ab{C}ut: markerless pose estimation of user-defined body parts with deep learning.
\newblock \emph{Nature Neuroscience}, pages 1281–--1289, 2018.

\bibitem[An et~al.(2023)An, Ren, Yu, Hai, Jia, and Liu]{MAMMAL}
Liang An, Jilong Ren, Tao Yu, Tang Hai, Yichang Jia, and Yebin Liu.
\newblock {Three-dimensional surface motion capture of multiple freely moving pigs using MAMMAL}.
\newblock \emph{Nature Communications}, 2023.

\bibitem[Belongie et~al.(2005)Belongie, Branson, Doll{\'a}r, and Rabaud]{smart_vivarium}
Serge Belongie, Kristin Branson, Piotr Doll{\'a}r, and Vincent Rabaud.
\newblock {Monitoring Animal Behavior in the Smart Vivarium}.
\newblock In \emph{Measuring Behavior}, pages 70--72, 2005.

\bibitem[Biderman et~al.(2024)Biderman, Whiteway, Hurwitz, Greenspan, Lee, Vishnubhotla, Warren, Pedraja, Noone, Schartner, Huntenburg, Khanal, Meijer, Noel, Pan-Vazquez, Socha, Urai, Laboratory, Cunningham, Sawtell, and Paninski]{lightningpose}
Dan Biderman, Matthew~R. Whiteway, Cole Hurwitz, Nicholas Greenspan, Robert~S. Lee, Ankit Vishnubhotla, Richard Warren, Federico Pedraja, Dillon Noone, Michael~M. Schartner, Julia~M. Huntenburg, Anup Khanal, Guido~T. Meijer, Jean-Paul Noel, Alejandro Pan-Vazquez, Karolina~Z. Socha, Anne~E. Urai, The International~Brain Laboratory, John~P. Cunningham, Nathaniel~B. Sawtell, and Liam Paninski.
\newblock {Lightning Pose: improved animal pose estimation via semi-supervised learning, Bayesian ensembling and cloud-native open-source tools}.
\newblock \emph{Nature Neuroscience}, pages 1316–--1328, 2024.

\bibitem[Bogo et~al.(2016)Bogo, Kanazawa, Lassner, Gehler, Romero, and Black]{smplify}
Federica Bogo, Angjoo Kanazawa, Christoph Lassner, Peter Gehler, Javier Romero, and Michael~J. Black.
\newblock {Keep it SMPL: Automatic Estimation of 3D Human Pose and Shape from a Single Image}.
\newblock In \emph{ECCV}, pages 561--578, 2016.

\bibitem[Bogo et~al.(2017)Bogo, Romero, Pons-Moll, and Black]{dynamicfaust}
Federica Bogo, Javier Romero, Gerard Pons-Moll, and Michael~J. Black.
\newblock Dynamic {FAUST}: {R}egistering human bodies in motion.
\newblock In \emph{CVPR}, pages 6233--6242, 2017.

\bibitem[Bohnslav et~al.(2023)Bohnslav, Osman, Jaggi, Soares, Weinreb, Datta, and Harvey]{Armo}
James~P Bohnslav, Mohammed Abdal~Monium Osman, Akshay Jaggi, Sofia Soares, Caleb Weinreb, Sandeep~Robert Datta, and Christopher~D Harvey.
\newblock {ArMo: An Articulated Mesh Approach for Mouse 3D Reconstruction}.
\newblock \emph{BioRxiv}, 2023.

\bibitem[Bolaños et~al.(2021)Bolaños, Xiao, Ford, LeDue, Gupta, Doebeli, Hu, Rhodin, and Murphy]{virtual_mouse}
Luis~A. Bolaños, Dongsheng Xiao, Nancy~L. Ford, Jeff~M. LeDue, Pankaj~K. Gupta, Carlos Doebeli, Hao Hu, Helge Rhodin, and Timothy~H. Murphy.
\newblock A three-dimensional virtual mouse generates synthetic training data for behavioral analysis.
\newblock \emph{Nature Methods}, 2021.

\bibitem[Butlera et~al.(2023)Butlera, Keim, Ray, and Azim]{color_label_rat}
Daniel~J. Butlera, Alexander~P. Keim, Shantanu Ray, and Eiman Azim.
\newblock Large-scale capture of hidden fluorescent labels for training generalizable markerless motion capture models.
\newblock \emph{Nature Communications}, 2023.

\bibitem[Chaumont et~al.(2012)Chaumont, Coura, Serreau, Cressant, Chabout, Granon, and Olivo-Marin]{mice_tracking}
Fabrice Chaumont, Renata Coura, Pierre Serreau, Arnaud Cressant, Jonathan Chabout, Sylvie Granon, and Jean-Christophe Olivo-Marin.
\newblock {Computerized video analysis of social interactions in mice}.
\newblock \emph{Nature Methods}, pages 410–--417, 2012.

\bibitem[Datta et~al.(2019)Datta, Anderson, Branson, Perona, and Leifer]{Computational_Neuroethology}
Sandeep~Robert Datta, David~J. Anderson, Kristin Branson, Pietro Perona, and Andrew Leifer.
\newblock {Computational Neuroethology: A Call to Action}.
\newblock \emph{Neuron}, pages 11--24, 2019.

\bibitem[Dell et~al.(2014)Dell, Bender, Branson, Couzin, {de Polavieja}, Noldus, Pérez-Escudero, Perona, Straw, Wikelski, and Brose]{Automated_imagebased}
Anthony~I. Dell, John~A. Bender, Kristin Branson, Iain~D. Couzin, Gonzalo~G. {de Polavieja}, Lucas~P.J.J. Noldus, Alfonso Pérez-Escudero, Pietro Perona, Andrew~D. Straw, Martin Wikelski, and Ulrich Brose.
\newblock {Automated image-based tracking and its application in ecology}.
\newblock \emph{Trends in Ecology and Evolution}, pages 417--428, 2014.

\bibitem[Deng et~al.(2022)Deng, Yao, Dyke, and Zhang]{non-rigid-registration-survey}
Bailin Deng, Yuxin Yao, Roberto~M. Dyke, and Juyong Zhang.
\newblock {A Survey of Non-Rigid 3D Registration}.
\newblock \emph{arXiv:2203.07858}, 2022.

\bibitem[Dunn et~al.(2021)Dunn, Marshall, Severson, Aldarondo, Hildebrand, Chettih, Wang, Gellis, Carlson, Aronov, et~al.]{dannce}
Timothy~W Dunn, Jesse~D Marshall, Kyle~S Severson, Diego~E Aldarondo, David~GC Hildebrand, Selmaan~N Chettih, William~L Wang, Amanda~J Gellis, David~E Carlson, Dmitriy Aronov, et~al.
\newblock {Geometric deep learning enables 3D kinematic profiling across species and environments}.
\newblock \emph{Nature Neuroscience}, pages 564--573, 2021.

\bibitem[Egnor and Branson(2016)]{computational_analysis}
S.E.~Roian Egnor and Kristin Branson.
\newblock {Computational Analysis of Behavior}.
\newblock \emph{Neuroscience and Biobehavioral Reviews}, pages 217--236, 2016.

\bibitem[Ge et~al.(2021)Ge, Liu, Wang, Li, and Sun]{yolox}
Zheng Ge, Songtao Liu, Feng Wang, Zeming Li, and Jian Sun.
\newblock {YOLOX: Exceeding YOLO Series in 2021}.
\newblock \emph{arXiv:2107.08430}, 2021.

\bibitem[Gosztolai et~al.(2021)Gosztolai, Günel, Abrate, Morales, Ríos, Rhodin, Fua, and Ramdya]{liftpose3d}
Adam Gosztolai, Semih Günel, Marco~Pietro Abrate, Daniel Morales, Victor Ríos, Helge Rhodin, Pascal Fua, and Pavan Ramdya.
\newblock {LiftPose3D, a deep learning-based approach for transforming 2D to 3D pose in laboratory animals}.
\newblock \emph{Nature Methods}, pages 975--981, 2021.

\bibitem[Goulding et~al.(2008)Goulding, Schenk, Juneja, MacKay, Wade, and Tecott]{A_robust_automated_system}
Evan~H. Goulding, A.~Katrin Schenk, Punita Juneja, Adrienne~W. MacKay, Jennifer~M. Wade, and Laurence~H. Tecott.
\newblock A robust automated system elucidates mouse home cage behavioral structure.
\newblock \emph{Proceedings of the National Academy of Sciences}, pages 20575--20582, 2008.

\bibitem[Graving et~al.(2019)Graving, Chae, Naik, Li, Koger, Costelloe, and Couzin]{deepposekit}
Jacob~M Graving, Daniel Chae, Hemal Naik, Liang Li, Benjamin Koger, Blair~R Costelloe, and Iain~D Couzin.
\newblock {DeepPoseKit, a software toolkit for fast and robust animal pose estimation using deep learning}.
\newblock \emph{eLife}, 2019.

\bibitem[Güler et~al.(2018)Güler, Neverova, and Kokkinos]{densepose}
Rıza~Alp Güler, Natalia Neverova, and Iasonas Kokkinos.
\newblock {DensePose: Dense Human Pose Estimation in the Wild}.
\newblock In \emph{CVPR}, pages 7297--7306, 2018.

\bibitem[Han et~al.(2020)Han, Chen, Liu, and Zwicker]{DRWR}
Zhizhong Han, Chao Chen, Yu-Shen Liu, and Matthias Zwicker.
\newblock {DRWR}: A differentiable renderer without rendering for unsupervised 3{D} structure learning from silhouette images.
\newblock In \emph{ICML}, pages 3994--4005, 2020.

\bibitem[Hartly and Zisserman(2004)]{hartly2004multiple}
Richard Hartly and Andrew Zisserman.
\newblock \emph{Multiple view geometry in computer vision}.
\newblock 2004.

\bibitem[Homberg and Alenina(2017)]{mice_vs_rat_3}
Judith~R. Homberg and Markus Wöhrand~Natalia Alenina.
\newblock {Comeback of the Rat in Biomedical Research}.
\newblock \emph{ACS Chemical Neuroscience}, 2017.

\bibitem[Hu et~al.(2023)Hu, Seybold, Yang, Sud, Liu, Barron, Cha, Cosino, Karlsson, Kite, Kolumam, Preciado, Zavala-Solorio, Zhang, Zhang, Voorbach, Tovcimak, Ruby, and Ross]{single_3D}
Bo Hu, Bryan Seybold, Shan Yang, Avneesh Sud, Yi Liu, Karla Barron, Paulyn Cha, Marcelo Cosino, Ellie Karlsson, Janessa Kite, Ganesh Kolumam, Joseph Preciado, José Zavala-Solorio, Chunlian Zhang, Xiaomeng Zhang, Martin Voorbach, Ann~E. Tovcimak, J.~Graham Ruby, and David~A. Ross.
\newblock {3D mouse pose from single-view video and a new dataset}.
\newblock \emph{Scientific Reports}, pages 2045--2322, 2023.

\bibitem[Huang et~al.(2021)Huang, Shu, Zhang, and Wang]{DMMR}
Buzhen Huang, Yuan Shu, Tianshu Zhang, and Yangang Wang.
\newblock {Dynamic Multi-Person Mesh Recovery From Uncalibrated Multi-View Cameras}.
\newblock In \emph{3DV}, pages 710--720. IEEE, 2021.

\bibitem[Huang et~al.(2017)Huang, Bogo, Lassner, Kanazawa, Gehler, Romero, Akhter, and Black]{MuvS}
Yinghao Huang, Federica Bogo, Christoph Lassner, Angjoo Kanazawa, Peter~V Gehler, Javier Romero, Ijaz Akhter, and Michael~J Black.
\newblock {Towards accurate marker-less human shape and pose estimation over time}.
\newblock In \emph{3DV}, pages 421--430, 2017.

\bibitem[Insafutdinov et~al.(2016)Insafutdinov, Pishchulin, Andres, Andriluka, and Schiele]{Deeper_cut}
Eldar Insafutdinov, Leonid Pishchulin, Bjoern Andres, Mykhaylo Andriluka, and Bernt Schiele.
\newblock {DeeperCut: A Deeper, Stronger, and Faster Multi-Person Pose Estimation Model}, 2016.

\bibitem[Joo et~al.(2015)Joo, Liu, Tan, Gui, Nabbe, Matthews, Kanade, Nobuhara, and Sheikh]{panopticstudio}
Hanbyul Joo, Hao Liu, Lei Tan, Lin Gui, Bart Nabbe, Iain Matthews, Takeo Kanade, Shohei Nobuhara, and Yaser Sheikh.
\newblock {Panoptic Studio: A Massively Multiview System for Social Motion Capture}.
\newblock In \emph{ICCV}, pages 3334--3342, 2015.

\bibitem[K et~al.(2019)K, M, W, and E]{mice_vs_rat_1}
Kondrakiewicz K, Kostecki M, Szadzińska W, and Knapska E.
\newblock Ecological validity of social interaction tests in rats and mice.
\newblock \emph{Genes Brain Behav}, 2019.

\bibitem[Kanazawa et~al.(2018)Kanazawa, Black, Jacobs, and Malik]{HMR}
Angjoo Kanazawa, Michael~J. Black, David~W. Jacobs, and Jitendra Malik.
\newblock {End-to-end Recovery of Human Shape and Pose}.
\newblock In \emph{CVPR}, 2018.

\bibitem[Kerbl et~al.(2023)Kerbl, Kopanas, Leimk{\"u}hler, and Drettakis]{3Dgaussiansplatting}
Bernhard Kerbl, Georgios Kopanas, Thomas Leimk{\"u}hler, and George Drettakis.
\newblock {3D Gaussian Splatting for Real-Time Radiance Field Rendering}.
\newblock \emph{ACM Transactions on Graphics}, 2023.

\bibitem[Kim et~al.(2023)Kim, Affan, Hadas~Frostig, and Alexander]{mice_vs_rat_2}
Su~Jin Kim, Rifqi~O. Affan, Benjamin B.~Scott Hadas~Frostig, and Andrew~S. Alexander.
\newblock Advances in cellular resolution microscopy for brain imaging in rats.
\newblock \emph{Neurophotonics}, 2023.

\bibitem[Kirillov et~al.(2023)Kirillov, Mintun, Ravi, Mao, Rolland, Gustafson, Xiao, Whitehead, Berg, Lo, Dollar, and Girshick]{SAM}
Alexander Kirillov, Eric Mintun, Nikhila Ravi, Hanzi Mao, Chloe Rolland, Laura Gustafson, Tete Xiao, Spencer Whitehead, Alexander~C. Berg, Wan-Yen Lo, Piotr Dollar, and Ross Girshick.
\newblock {Segment Anything}.
\newblock In \emph{ICCV}, pages 4015--4026, 2023.

\bibitem[Kolotouros et~al.(2019)Kolotouros, Pavlakos, Black, and Daniilidis]{SPIN}
Nikos Kolotouros, Georgios Pavlakos, Michael~J Black, and Kostas Daniilidis.
\newblock {Learning to reconstruct 3D human pose and shape via model-fitting in the loop}.
\newblock In \emph{ICCV}, pages 2252--2261, 2019.

\bibitem[Kuhn(1955)]{hungarian}
Harold~W Kuhn.
\newblock {The Hungarian method for the assignment problem}.
\newblock \emph{Naval research logistics quarterly}, pages 83--97, 1955.

\bibitem[L{\'e}vy(2006)]{levy2006laplace}
Bruno L{\'e}vy.
\newblock Laplace-beltrami eigenfunctions towards an algorithm that" understands" geometry.
\newblock In \emph{IEEE International Conference on Shape Modeling and Applications}, pages 13--13, 2006.

\bibitem[Li et~al.(2017)Li, Bolkart, Black, Li, and Romero]{FLAME:SiggraphAsia2017}
Tianye Li, Timo Bolkart, Michael.~J. Black, Hao Li, and Javier Romero.
\newblock Learning a model of facial shape and expression from {4D} scans.
\newblock \emph{ACM Transactions on Graphics}, pages 194:1--194:17, 2017.

\bibitem[{L}oper et~al.(2015){L}oper, {M}ahmood, {R}omero, {P}ons {M}oll, and {B}lack]{SMPL}
{M}atthew {L}oper, {N}aureen {M}ahmood, {J}avier {R}omero, {G}erard {P}ons {M}oll, and {M}ichael~{J}. {B}lack.
\newblock {SMPL}: {A} {S}kinned {M}ulti-{P}erson {L}inear {M}odel.
\newblock \emph{ACM Transactions on Graphics}, pages 248:1--248:16, 2015.

\bibitem[Lorbach et~al.(2018)Lorbach, Kyriakou, Poppe, {van Dam}, Noldus, and Veltkamp]{learning_to_recognize}
Malte Lorbach, Elisavet~I. Kyriakou, Ronald Poppe, Elsbeth~A. {van Dam}, Lucas~P.J.J. Noldus, and Remco~C. Veltkamp.
\newblock {Learning to recognize rat social behavior: Novel dataset and cross-dataset application}.
\newblock \emph{Journal of Neuroscience Methods}, pages 166--172, 2018.

\bibitem[Luiten et~al.(2024)Luiten, Kopanas, Leibe, and Ramanan]{dynamic3dgaussian}
Jonathon Luiten, Georgios Kopanas, Bastian Leibe, and Deva Ramanan.
\newblock {Dynamic 3D Gaussians: Tracking by persistent dynamic view synthesis}.
\newblock In \emph{3DV}, pages 800--809, 2024.

\bibitem[Maghsoudi et~al.(2017)Maghsoudi, Tabrizi, Robertson, , and Spence]{superpixcel}
Omid~Haji Maghsoudi, Annie~Vahedipour Tabrizi, Benjamin Robertson, , and Andrew Spence.
\newblock {Superpixels Based Marker Tracking Vs. Hue Thresholding In Rodent Biomechanics Application}.
\newblock \emph{arXiv:1710.06473}, 2017.

\bibitem[Marshall et~al.(2021)Marshall, Aldarondo, Dunn, Wang, Berman, and Ölveczky]{Continuous_Whole_Body_3D}
Jesse~D. Marshall, Diego~E. Aldarondo, Timothy~W. Dunn, William~L. Wang, Gordon~J. Berman, and Bence~P. Ölveczky.
\newblock {Continuous Whole-Body 3D Kinematic Recordings across the Rodent Behavioral Repertoire}.
\newblock \emph{Neuron}, pages 420--437, 2021.

\bibitem[Martinez et~al.(2017)Martinez, Hossain, Romero, and Little]{humanliftpose3d}
Julieta Martinez, Rayat Hossain, Javier Romero, and James~J. Little.
\newblock {A Simple yet Effective Baseline for 3D Human Pose Estimation}.
\newblock In \emph{ICCV}, pages 2640--2649, 2017.

\bibitem[Matsumoto et~al.(2013)Matsumoto, Urakawa, Takamura, Malcher-Lopes, Hori, Tomaz, Ono, and Nishijo]{3d_rat_recovery}
Jumpei Matsumoto, Susumu Urakawa, Yusaku Takamura, Renato Malcher-Lopes, Etsuro Hori, Carlos Tomaz, Taketoshi Ono, and Hisao Nishijo.
\newblock {A 3D-Video-Based Computerized Analysis of Social and Sexual Interactions in Rats}.
\newblock \emph{PLoS One}, page e78460, 2013.

\bibitem[Mimica et~al.(2018)Mimica, Dunn, Tombaz, Bojja, and Whitlock]{mocap_rat}
Bartul Mimica, Benjamin~A. Dunn, Tuce Tombaz, V.~P. T. N. C.~Srikanth Bojja, and Jonathan~R. Whitlock.
\newblock {E}fficient cortical coding of {3D} posture in freely behaving rats.
\newblock \emph{Science}, pages 584--589, 2018.

\bibitem[Nagy et~al.(2024)Nagy, Davidson, Vasarhelyi, Abel, Kubinyi, Hady, and Vicsek]{long_term_tracking_socialstructure}
Mate Nagy, Jacob~D. Davidson, Gabor Vasarhelyi, Daniel Abel, Eniko Kubinyi, Ahmed~El Hady, and Tamas Vicsek.
\newblock {Long-term tracking of social structure in groups of rats}.
\newblock \emph{arXiv:2408.08945}, 2024.

\bibitem[Neverova et~al.(2020)Neverova, Novotny, Szafraniec, Khalidov, Labatut, and Vedaldi]{CSE}
Natalia Neverova, David Novotny, Marc Szafraniec, Vasil Khalidov, Patrick Labatut, and Andrea Vedaldi.
\newblock {Continuous Surface Embeddings}.
\newblock In \emph{NeurIPS}, pages 17258--17270, 2020.

\bibitem[Pereira et~al.(2019)Pereira, Aldarondo, Willmore, Kislin, Wang, Murthy, and Shaevitz]{LEEP}
T.~D. Pereira, D.~E. Aldarondo, L. Willmore, M. Kislin, S.~S. Wang, M. Murthy, and J.~W Shaevitz.
\newblock {Fast animal pose estimation using deep neural networks}.
\newblock \emph{Nature Neuroscience}, pages 117--125, 2019.

\bibitem[Qian et~al.(2024)Qian, Kirschstein, Schoneveld, Davoli, Giebenhain, and Nie{\ss}ner]{gaussianavatars}
Shenhan Qian, Tobias Kirschstein, Liam Schoneveld, Davide Davoli, Simon Giebenhain, and Matthias Nie{\ss}ner.
\newblock {GaussianAvatars: Photorealistic Head Avatars with Rigged 3D Gaussians}.
\newblock In \emph{CVPR}, pages 20299--20309, 2024.

\bibitem[Ravi et~al.(2024)Ravi, Gabeur, Hu, Hu, Ryali, Ma, Khedr, Rädle, Rolland, Gustafson, Mintun, Pan, Alwala, Carion, Wu, Girshick, Dollár, and Feichtenhofer]{SAM2}
Nikhila Ravi, Valentin Gabeur, Yuan-Ting Hu, Ronghang Hu, Chaitanya Ryali, Tengyu Ma, Haitham Khedr, Roman Rädle, Chloe Rolland, Laura Gustafson, Eric Mintun, Junting Pan, Kalyan~Vasudev Alwala, Nicolas Carion, Chao-Yuan Wu, Ross Girshick, Piotr Dollár, and Christoph Feichtenhofer.
\newblock {SAM 2: Segment Anything in Images and Videos}.
\newblock \emph{arXiv:2408.00714}, 2024.

\bibitem[Redmon and Farhadi(2024)]{psavatar}
Joseph Redmon and Ali Farhadi.
\newblock {PSA}vatar: {A} point-based {S}hape {M}odel for {R}eal-{T}ime {H}ead {A}vatar {A}nimation with {3D} {G}aussian {S}platting.
\newblock \emph{arXiv:2401.12900}, 2024.

\bibitem[{R}emy {S}abathier et~al.(2024){R}emy {S}abathier, {N}iloy~{J}yoti {M}itra, and {D}avid {N}ovotny]{animalavatars}
{R}emy {S}abathier, {N}iloy~{J}yoti {M}itra, and {D}avid {N}ovotny.
\newblock {A}nimal {A}vatars: {R}econstructing {A}nimatable {3D} {A}nimals from {C}asual {V}ideos.
\newblock \emph{arXiv:2403.17103}, 2024.

\bibitem[Sanakoyeu et~al.(2020)Sanakoyeu, Khalidov, McCarthy, Vedaldi, and Neverova]{DensePose_chimps}
Artsiom Sanakoyeu, Vasil Khalidov, Maureen~S. McCarthy, Andrea Vedaldi, and Natalia Neverova.
\newblock {Transferring Dense Pose to Proximal Animal Classes}.
\newblock In \emph{CVPR}, pages 5233--5242, 2020.

\bibitem[Shao et~al.(2024)Shao, Wang, Li, Wang, Lin, Zhang, Fan, and Wang]{SplattingAvatar}
Zhijing Shao, Zhaolong Wang, Zhuang Li, Duotun Wang, Xiangru Lin, Yu Zhang, Mingming Fan, and Zeyu Wang.
\newblock {SplattingAvatar: Realistic Real-Time Human Avatars with Mesh-Embedded Gaussian Splatting}.
\newblock In \emph{CVPR}, pages 1606--1616, 2024.

\bibitem[Sorkine and Alexa(2007)]{arap}
Olga Sorkine and Marc Alexa.
\newblock {As-Rigid-As-Possible Surface Modeling}.
\newblock In \emph{Symposium on Geometry processing}, pages 109--116, 2007.

\bibitem[Vaswani et~al.(2017)Vaswani, Shazeer, Parmar, Uszkoreit, Jones, Gomez, Kaiser, and Polosukhin]{transformer}
Ashish Vaswani, Noam Shazeer, Niki Parmar, Jakob Uszkoreit, Llion Jones, Aidan~N Gomez, \L~ukasz Kaiser, and Illia Polosukhin.
\newblock {Attention is All you Need}.
\newblock In \emph{NeurIPS}, pages 5998--6008, 2017.

\bibitem[Ye et~al.(2024)Ye, Filippova, Lauer, Schneider, Vidal, Qiu, Mathis, and Mathis]{ye2024superanimal}
Shaokai Ye, Anastasiia Filippova, Jessy Lauer, Steffen Schneider, Maxime Vidal, Tian Qiu, Alexander Mathis, and Mackenzie~Weygandt Mathis.
\newblock Superanimal pretrained pose estimation models for behavioral analysis.
\newblock \emph{Nature Communications}, page 5165, 2024.

\bibitem[Youwang et~al.(2021)Youwang, Ji-Yeon, Joo, and Oh]{Youwang2021Unified3M}
Kim Youwang, Kim Ji-Yeon, Kyungdon Joo, and Tae-Hyun Oh.
\newblock {Unified 3D Mesh Recovery of Humans and Animals by Learning Animal Exercise}.
\newblock \emph{BMVC}, 2021.

\bibitem[Zimmermann et~al.(2020)Zimmermann, Schneider, Alyahyay, Brox, and Diester]{Freipose2020}
Christian Zimmermann, Artur Schneider, Mansour Alyahyay, Thomas Brox, and Ilka Diester.
\newblock {FreiPose: a deep learning framework for precise animal motion capture in 3D spaces}.
\newblock \emph{BioRxiv}, pages 2020--02, 2020.

\bibitem[Zimmermann et~al.(2023)Zimmermann, Schneider, Alyahyay, Brox, and Diester]{AMBER}
Christian Zimmermann, Artur Schneider, Mansour Alyahyay, Thomas Brox, and Ilka Diester.
\newblock {Automated maternal behavior during early life in rodents (AMBER) pipeline}.
\newblock \emph{Scientific Reports}, 2023.

\bibitem[{Z}uffi et~al.(2017){Z}uffi, {K}anazawa, {J}acobs, and {B}lack]{SMAL}
{S}ilvia {Z}uffi, {A}ngjoo {K}anazawa, {D}avid~{W}. {J}acobs, and {M}ichael~{J}. {B}lack.
\newblock 3{D} {M}enagerie: {M}odeling the {3D} {S}hape and {P}ose of {A}nimals.
\newblock In \emph{CVPR}, pages 6365--6373, 2017.

\bibitem[{Z}uffi et~al.(2024){Z}uffi, {M}ellbin, {L}i, {H}oeschle, {K}jellstr\"om, {P}olikovsky, {H}ernlund, and {B}lack]{VAREN}
{S}ilvia {Z}uffi, {Y}lva {M}ellbin, {C}i {L}i, {M}arkus {H}oeschle, {H}edvig {K}jellstr\"om, {S}enya {P}olikovsky, {E}lin {H}ernlund, and {M}ichael~{J}. {B}lack.
\newblock {VAREN}: {V}ery {A}ccurate and {R}ealistic {E}quine {N}etwork.
\newblock In \emph{CVPR}, pages 5374--5383, 2024.

\end{thebibliography}
}

\end{document}